\documentclass[journal]{IEEEtran}
\usepackage[square, comma, sort&compress, numbers]{natbib}
\usepackage[linesnumbered,ruled,vlined]{algorithm2e}
\usepackage{caption}
\usepackage{graphicx, subfig}
\usepackage{hyperref}
\usepackage{multirow}
\usepackage{epsfig}
\usepackage{amsmath}
\usepackage{indentfirst}
\usepackage{amssymb}
\usepackage{lineno,hyperref}
\usepackage{setspace}
\usepackage{amsfonts}
\usepackage{url}
\usepackage{enumitem}
\usepackage{algorithmic}
\usepackage{enumerate}
\usepackage{longtable}
\usepackage{booktabs}
\usepackage{array}
\usepackage[figuresright]{rotating}
\usepackage{lscape}
\usepackage{amssymb}
\usepackage{tabulary}
\usepackage{color}
\usepackage{longtable}
\makeatletter

\newcommand{\Rmnum}[1]{\expandafter\@slowromancap\romannumeral #1@}
\makeatother
\hypersetup
{
	pdfauthor = UnnamedOrange,
	colorlinks = true,
	linkcolor = black,
	anchorcolor = black,
	citecolor = black,
	urlcolor = black
}

\usepackage{float}
\ifCLASSINFOpdf
 
\else

\fi

\begin{document}

\title{Overview and Experimental Study of Learning-based Optimization Algorithms for the Vehicle Routing Problem}

\author{Bingjie Li, Guohua Wu, Yongming He, Mingfeng Fan, and Witold Pedrycz~\IEEEmembership{Fellow,~IEEE,}
\thanks{Bingjie Li Guohua Wu and Mingfeng Fan are with the School of Traffic and Transportation Engineering, Central South University, Changsha 410075, China. E-mail: csulbj@csu.edu.cn; guohuawu@csu.edu.cn;mingfan@csu.edu.cn.}
\thanks{Yongming He is with the College of Systems Engineering, National University of Defense Technology, Changsha 410073, China (e-mail: heyongming10@hotmail.com)}
\thanks{Witold Pedrycz is with the Department of Electrical and Computer Engineering, University of Alberta, Edmonton, AB T6G 2V4, Canada, with the Department of Electrical and Computer Engineering, Faculty of Engineering, King Abdulaziz University, Jeddah 21589, Saudi Arabia, and also with the Systems Research Institute, Polish Academy of Sciences, Warsaw 01447, Poland. E-mail: wpedrycz@ualberta.ca.}
}

\maketitle

\begin{abstract}
Vehicle routing problem (VRP) is a typical discrete combinatorial optimization problem, and many models and algorithms have been proposed to solve the VRP and its variants. Although existing approaches have contributed a lot to the development of this field, these approaches either are limited in problem size or need manual intervening in choosing parameters. To solve these difficulties, many studies have considered the learning-based optimization (LBO) algorithms to solve the VRP. This paper reviews recent advances in this field and divides relevant approaches into \emph{end-to-end approaches} and \emph{step-by-step approaches}. We performed a statistical analysis of the reviewed articles from various aspects and designed three experiments to evaluate the performance of four representative LBO algorithms. Finally, we conclude the applicable types of problems for different LBO algorithms and suggest directions in which researchers can improve LBO algorithms.
\end{abstract}


%
\IEEEpeerreviewmaketitle

\section{Introduction}
%
%
%
%
\IEEEPARstart{T}{he} VRP is one of the most widely researched problems in transportation and operations research areas \cite{ref1}. The canonical VRP has a simple structure and can be seen as a basic discrete combinatorial optimization problem. Many optimization problems can be transformed into the form of the VRP. Hence, the VRP can be used to demonstrate the optimization performance of different algorithms, which can help to design efficient algorithms. Moreover, the VRP has significant practicality in the real world. VRP variants have appeared in many fields \cite{ref6,ref7}, particularly in the logistic industry that is flourishing with the development of the globalization. Effective routing optimization can save a lot of cost (generally ranging from 5\% to 20\%) \cite{ref8}, which is of great significance for improving distribution efficiency \cite{ref1,ref9,ref10,ref11,ref12} and increasing economic benefits of enterprises \cite{ref13,ref14,ref15}. The VRP is still continuously attracting attention from researchers \cite{ref4, ref23}. However, it is still an challenging optimization task due to the NP-hard characteristic \cite{ref24} and its different complex variants.

\begin{figure}[htb]
	\begin{center}
		\includegraphics[width=3.0in,height=1.4in]{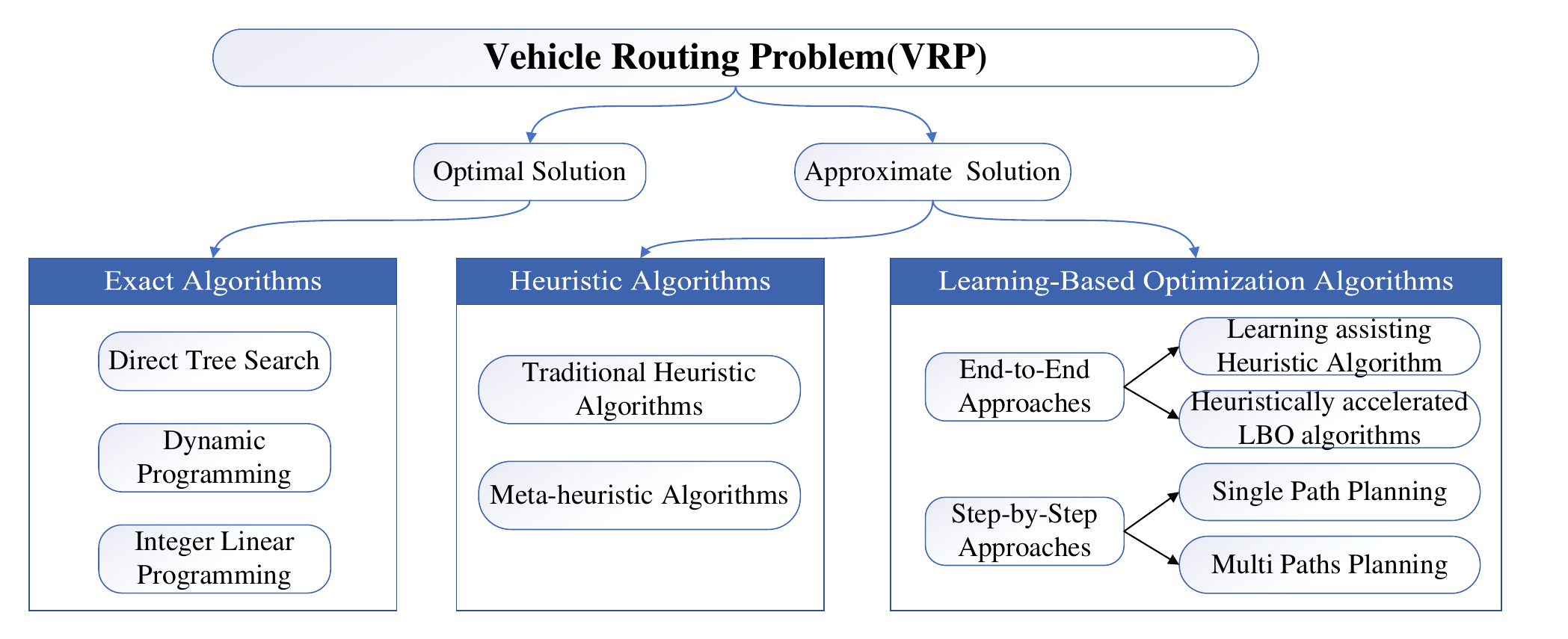}
	\end{center}
	\caption{Overall classification of the referred algorithms of the VRP.}\label{Fig:3}
\end{figure}

Optimization algorithms for solving the VRP can be crudely divided into three categories: exact, heuristic, and the LBO algorithms. Fig.~\ref{Fig:3} shows the overview of referred algorithms. Heuristic algorithms and exact algorithms as traditional algorithms for combinatorial optimization problems, many literatures have reviewed their applications in the VRP \cite{refgolden}. Ma{\'n}dziuk \cite{ref18} reviewed papers of using heuristic algorithms to solve the VRP between 2015 and 2017. Adewumi and Adeleke \cite{refadewumi} emphasized recent advances of traditional algorithms in 2018, while Dixit \emph{et al.} \cite{refdixit} reviewed some of the recent advancements of meta-heuristic techniques in solving the VRP with time windows (VRPTW). Exact and heuristic algorithms have made significant progress in the VRP, but the recent research trend is to design a flexible model that can solve the VRP with large scale and complex constraints more quickly. Exact algorithms need adequate time to get an optimal solution when solving a VRP with large scale \cite{ref21}. The optimality of heuristic algorithms cannot be guaranteed and their computational complexity is not satisfactory \cite{refgoel}. What's more, these two kinds of algorithms usually need specific design for solving the concrete problem. In recent years, the LBO algorithms have been applied to different combinatorial optimization problems \cite{ref66,ref67}, and many remarkable research breakthroughs have been achieved. Hence, the LBO algorithms have attracted significant attention in the field of the VRP.

The LBO algorithms can learn a model from training sets to obtain optimization strategies \cite{ref42,ref43}, which could automatically produce solutions of online tasks by end-to-end or step-by-step approaches. In end-to-end approaches, the model is trained to approximate a mapping function between the input and the solution, and the model can directly output a feasible solution when given an unknown task in application. While step-by-step approaches learn optimization strategies that could iteratively improve a solution rather than outputting a final solution directly. Both approaches have strong learning ability and generalization. They can overcome the deficiencies of the tedious parameter tuning of exact and heuristic algorithms and rapidly solve online instances through the advantage of offline training.  However, the LBO algorithms still have some technological bottlenecks to be settled, such as training data limitation, generality limitation, and so on.

Overall, although the application of the LBO algorithms to the VRP has become popular and significant achievements have been gotten \cite{ref40,ref41} in recent years, up to now there are no comprehensive assessment experiments and in-depth analyses on the characteristics of different LBO algorithms in solving the VRP. Therefore, we aim to present an elaborate overview and experimental study of related work to fill the gap in this field. Contributions of the paper are mainly on the following aspects:
	\begin{itemize}
		\item The paper briefly introduces the applications of the LBO algorithms to the VRP to aid beginners in understanding the development of this field.
		\item The paper discusses the advantages and disadvantages of different LBO algorithms based on extensive experiments on different datasets.
		\item The paper suggests promising research trends on using the LBO algorithms to solve the VRP in the future.
\end{itemize}

The remainder of this paper is structured as follows. In Section II, we introduce the background of the VRP, including the mathematical model and classical optimization algorithms. In Section III, we review the related literatures and provide a summative analysis of the references. In Section IV, the experimental comparisons among different algorithms are presented. The final section summarizes the full text and suggests several research trends of using the LBO algorithms to solve the VRP.
\vspace{-0.5cm}
\vspace{-0.5cm}
\section{Background of the VRP}
As the routing optimization problems are increasingly crucial in industrial engineering, logistics, etc, a series of research breakthroughs have been achieved in the VRP and many algorithms have been proposed. In this section, we briefly introduce the background of the VRP.
\subsection{Variants of the VRP}
\begin{figure}[htb]
	\begin{center}
		\includegraphics[width=3.3in]{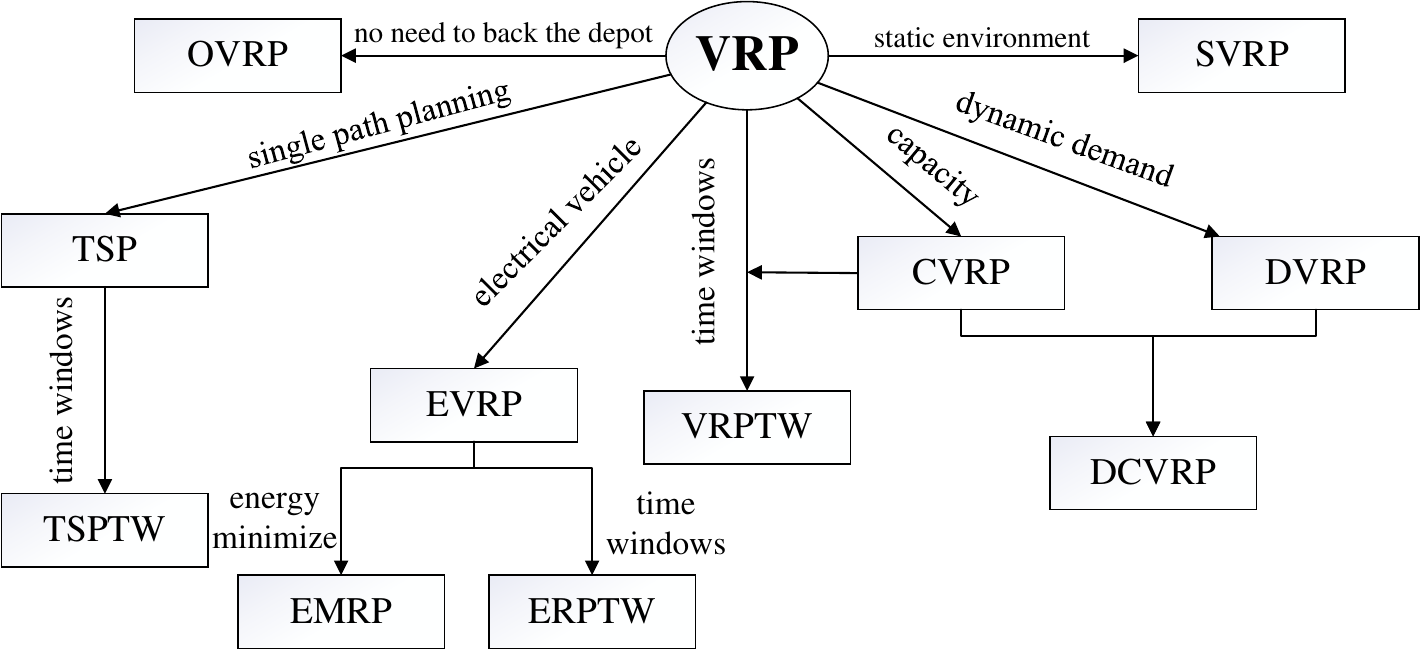}
	\end{center}
	\caption{Different variants of the VRP.}\label{Fig:1}	
\end{figure}

The mathematical model of the VRP is first proposed by Dantzig and Ramser \cite{ref2} in 1959. A few years later, Clarke and Wright \cite{ref3} proposed an effective greedy algorithm to solve the VRP, which also launched a research boom in the VRP. The VRP can be defined as follows: for a series of nodes with different demands, the aim is to plan routes with the lowest total cost under various side constraints \cite{ref16,ref17} for vehicles to serve these nodes sequentially according to the planning routes. With its wide application in the real world, the VRP must consider more constraints, which results in the derivation of many variants. Fig.~\ref{Fig:1} presents the hierarchy of VRP variants.

\begin{figure}[htb]
	\begin{center}
		\includegraphics[width=2.0in]{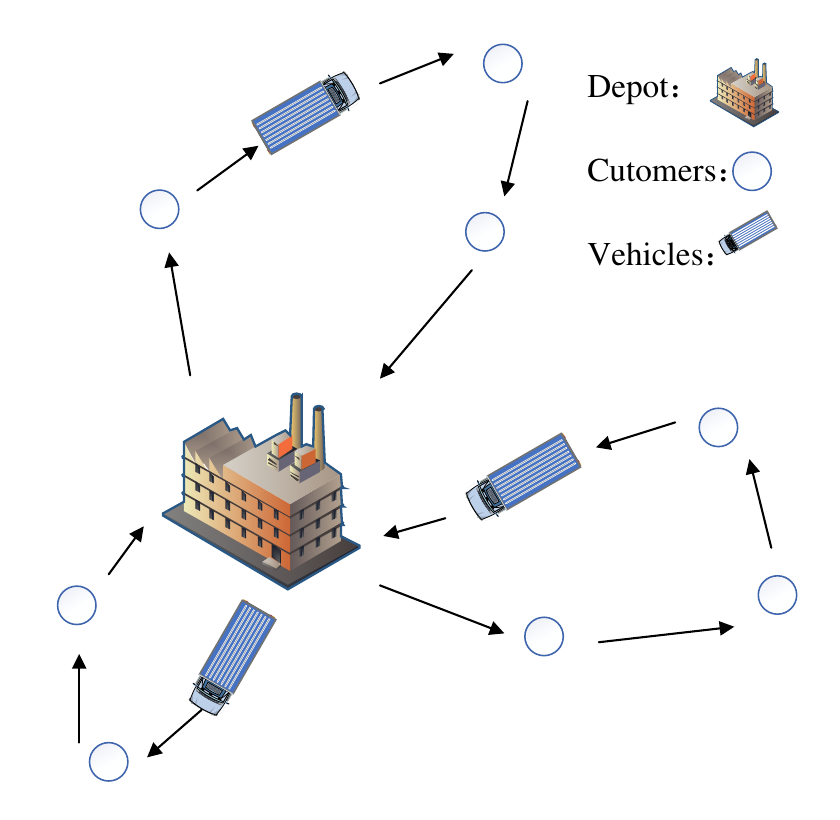}
	\end{center}
	\caption{Schematic diagram of the CVRP}\label{Fig:2}
\end{figure}

The basic version of the VRP is the capacitated vehicle routing problem (CVRP)  \cite{ref18,ref19}. In the CVRP, each customer is served only once and the sum of goods carried by a vehicle should not exceed its capacity \cite{ref20}. This type of problem has the following characteristics: (1) vehicles leave from the depot and return to it; (2) vehicles are permitted to visit each customer on a set of routes exactly once; (3) each customer has a non-negative demand of delivery. Fig.~\ref{Fig:2} is the schematic diagram of the CVRP.
\vspace{-1cm}

~\
\subsection{Mathematical model of the VRP}
The VRP can be described by an integer programming model \cite{ref21}, and we mainly present the mathematical model of the CVRP in this subsection. According to \cite{ref22}, the mathematical model of the CVRP can be expressed as follows, where \emph{N} represents the set of customers, the number 0 represents the depot, \emph{K} represents the number of vehicles, \emph{A} represents the set of arcs and \emph{V} represents the set of nodes, with $V=N\cup\{0\}$.  We define a binary decision variable \emph{x$_{i,j}^k$} to be 1 if the pair of nodes \emph{i} and \emph{j} are adjacent in the route of the vehicle \emph{$k$}; otherwise, \emph{x$_{i,j}^k$} equals 0. Each vehicle's maximum capacity is \emph{C}. \emph{u$_i^k$}  represents the surplus capacity of the vehicle \emph{k} after it serves the customer \emph{i}, and \emph{q$_j$} indicates the demand of customer \emph{j}.

\begin{small}
\begin{equation}\label{Eq:1}
	{\rm{minimize}}{\kern 1pt} {\kern 1pt} {\kern 1pt} {\kern 1pt} {\kern 1pt} {\kern 1pt} {\kern 1pt} {\kern 1pt} {\kern 1pt} {\kern 1pt} {\kern 1pt} \sum_{k\in K}\sum_{i,j\in A} c_{i,j}x_{i,j}^{k},
\end{equation}
\begin{equation}\label{Eq:2}
	{\rm{S.t.}}  \sum_{k\in K}\sum_{j\in V,j\ne i}x_{i,j}^{k}= 1, i\in N
\end{equation}
\begin{equation}\label{Eq:3}
	{\kern 1pt} {\kern 1pt}\sum_{k\in K}\sum_{i\in V,i\ne j} x_{i,j}^{k}= 1, j\in N
\end{equation}	
\begin{equation}\label{Eq:4}
	\sum_{k\in K}\sum_{i\in N} x_{i,0}^{k}= K
\end{equation}
\begin{equation}\label{Eq:5}
	\sum_{k\in K}\sum_{j\in N} x_{0,j}^{k}= K
\end{equation}
\begin{equation}\label{Eq:6}
		u_j^{k}=
		\begin{cases}
			u_i^{k}-q_j,\: \:  &if {\kern 1pt} x_{i,j}^{k}=1, \ \{i,j\} \in A: j\ne 0\\
		\emph{C}, \: \: \: \: \: & if \ j=0
	\end{cases}
\end{equation}
\begin{equation}\label{Eq:7}
	q_j \le u_i^{k} \le C, if \ x_{i,j}^{k}=1: i \in V, \ j \in N 
\end{equation}
\begin{equation}\label{Eq:8}
	x_{i,j}^{k} \in \{0,1\}, \{i,j\}\in A
\end{equation}
	\end{small}


In this formulation, the objective is to minimize the total cost of transportation and \emph{c$_{i,j}$} represents the cost of arc \{\emph{i,j}\}. Equations (\ref{Eq:2}) and  (\ref{Eq:3}) indicate that each customer can be served only once. Constraints (\ref{Eq:4}) and (\ref{Eq:5}) represent that both number of vehicles leaving the depot and returning to the depot should be equal to the total number of vehicles, which means that each vehicle should start from and end at the depot. Equation (\ref{Eq:6}) shows the update process of the surplus capacity of the vehicle \emph{k}, and constraint (\ref{Eq:7}) represents that the surplus capacity of the vehicle must not be less than the demand of its next customer \emph{j} while does not exceed the maximum capacity \emph{C}. Constraint (\ref{Eq:8}) is used to ensure decision variable \emph{x$_{i,j}^k$} is a binary.
\subsection{Traditional Optimization Algorithms for the VRP}
 Traditional optimization algorithms for solving the VRP include exact algorithms and heuristic algorithms. Fisher \cite{ref25} divided the development of traditional algorithms applied to the VRP into three stages. From 1959 to 1970 is the first stage, and simple heuristic algorithms were mainly used in this stage, such as the local heuristic and greedy algorithms; the second stage is from 1970 to 1980, and this stage primarily applied exact algorithms; after 1980 is the third stage and meta-heuristic algorithms began to be applied to VRPs. These traditional algorithms have been successfully used to solve different VRP variants\cite{refcaceres}.

\subsubsection{Exact algorithms}
According to \cite{ref26}, exact algorithms for solving the VRP can be divided into three categories: direct tree search (such as the branch-and-bound algorithm \cite{ref27}), dynamic programming \cite{ref28}, and integer linear programming (such as the column generation algorithm \cite{ref29}). Many breakthroughs were achieved using exact algorithms in the 1970s. Christofides and Eilon \cite{ref30} studied the CVRP and dynamic VRP (DVRP) by designing three approaches: a branch-and-bound approach, a saving approach, and a 3-optimal tour method. Christofides \emph{et al.} \cite{ref31}  used tree search algorithms to solve the VRP. Although these exact algorithms are easy to understand, their computational expense is prohibitive in large-scale problems \cite{ref32,ref33}.
\subsubsection{Heuristic algorithms}
Heuristic algorithms can be divided into traditional heuristic algorithms and meta-heuristic algorithms. Referring to \cite{ref34}, traditional heuristic algorithms can be divided into constructive and two-stage heuristic algorithms. The former gradually generates feasible solutions to minimize costs (e.g. the savings algorithm \cite{ref3}). The latter first clusters nodes and then constructs multiple feasible routes to satisfy constraints. The quality of the solution is improved by changing the positions of nodes between or within routes (e.g. \cite{ref35}). Constructing the neighborhood of a solution is a significant part of heuristic algorithms. However, traditional heuristic algorithms perform an entire search without focus; therefore, they only explore neighborhoods in a limited manner. A meta-heuristic algorithm can focus on potential areas by combining intelligently complex search rules \cite{ref36} and memory structures. As reported in \cite{ref37}, meta-heuristic algorithms generally produce better solutions than traditional heuristic algorithms (typically between 3\% and 7\%).

Although exact and heuristic algorithms have been developed for many years, they all have their own limitations. Both algorithms frequently require optimization expertise to model the problem and design effective search rules. Moreover, algorithms based on search frequently encounter the challenge of a trade-off between searching efficiency and solution accuracy when solving complex combinatorial optimization problems. This is apparent because finding an optimal solution comes at the expense of searching for a larger neighborhood and longer computational time. With the development of computer technology since 2010, the LBO algorithms have become a popular research topic and many breakthroughs have been obtained in using them to solve optimization problems. Along with the deepening of research on the VRP, it is necessary to design algorithms that can more rapidly and efficiently solve problems. Hence, the LBO algorithms are beginning to be applied to solve the VRP.

\section{Learning-based Optimization Algorithms \\for the VRP}\label{sec3}
From a technical point of view, the LBO algorithms usually include three kinds of learning modes \cite{ref46,ref47}. If the agent learns on data with labels, this training mode is called supervised learning (SL). In contrast, learning from unlabeled data is named unsupervised learning (UL). Note that UL is commonly used for the parameter optimization of continuous problems; therefore, it is rarely used in the literatures reviewed in our paper. The final type of learning framework is reinforcement learning (RL), which requires the agent to learn from interaction with the environment. RL involves agents sense the environment and learn to select optimal actions through trial and error \cite{ref48,ref49}. Compared with traditional optimization algorithms, the LBO algorithms have three advantages:
\begin{itemize}
	\item The LBO algorithms do not require substantial domain knowledge \cite{ref39} for mathematical modeling and parameter tuning, which enables the LBO algorithms to model routing problems in a real-world more flexiblely. 
	\item The LBO algorithms can automatically construct an empirical formula to approximate the mapping function between the inputs and solutions through pre-training on a dataset.
	\item The LBO algorithms can automatedly extract optimization knowledge from training data and give computers the ability to learn without being explicitly programmed \cite{ref38}. 
\end{itemize}

The ability of autonomous and offline learning results in LBO models not requiring any hand-engineered reasoning and rapidly providing a promising solution for unknown data \cite{ref39}. The LBO algorithms have been applied to different fields \cite{ref43}, such as video games \cite{ref79}, Go \cite{ref80}, robotics control \cite{ref81} and image identification \cite{ref82}. Many studies have used the LBO algorithms for the VRP, and we divide related frameworks into two types: end-to-end and step-by-step approaches.
\subsection{Step-by-step approaches}
Step-by-step approaches learn optimization strategies that could iteratively improve a solution rather than outputting a final solution directly. These approaches can find a promising solution, but they have low time efficiency for training since their search space is more extensive. We subdivide this type of approaches into learning assisting heuristic and heuristically accelerated LBO algorithms according to different solving frameworks.
\subsubsection{Learning assisting heuristic algorithms}Heuristic algorithms use a series of operators to find the optimal solution \cite{ref44}. During search process, heuristic algorithms would generate considerable information about how to evolve and search in different stages \cite{ref83}, but this useful information is not comprehensively and well utilized by heuristic algorithms. Hence, many studies have used the LBO algorithms as assistors to learn from this information and define the optimal setting of heuristic algorithms. Hence, we call algorithms of these studies \emph{learning assisting heuristic algorithms} because they still search based on heuristic framework.

In their book in 1995, Gambardella \emph{et al.} \cite{ref84} used Q-learning to construct the initial path for the ant system (AS). The LBO models in both papers of Lima \emph{et al.} \cite{ref85} and Alipour  \emph{et al.} \cite{ref86} were also used as constructors of initial solutions for heuristic algorithms. Subsequently, many scholars used the LBO models to assist heuristic algorithms from different aspects. Liu \emph{et al.} \cite{ref87} proposed an improved generic algorithm (GA) with RL called RMGA. They used RL model to determine whether to break the connection relation of the initial tour and used the GA to reselect the next city for the current city. Later, Phiboonbanakit \emph{et al.} \cite{ref88} used a transfer learning algorithm to aid the GA in dividing customers into regional clusters beforehand. Ding \emph{et al.} \cite{ref89} used the LBO models to predict the values of nodes for accelerating convergence of solvers, and recently, Sun \emph{et al.} \cite{ref90} used the LBO model to quantify the likelihood of each edge belonging to an optimal route by the support vector machine (SVM). The LBO models can reduce the search space to simplify problems for heuristics, and all experiments demonstrated that these learning assisting heuristic approaches can significantly improve the performance of original heuristic algorithms and speed up the solving process.

In addition to using the LBO algorithms to assist heuristic algorithms during constructing solution, the LBO algorithms can set parameters of heuristic algorithms. In 2017, Cooray and Rupasinghe \cite{ref91} used the LBO model to set the mutation rate of the GA for the energy-minimized VRP (EMVRP). They experimented with GA with different mutation rates on instances and selected the optimal parameter settings that can produce minimal energy consumption. It has been proven that parameter tuning according to data characteristics has a significant effect on improving the applicability of the GA. Al-Duoli \emph{et al.} \cite{ref92} also used the LBO model to define the parameters values of a heuristic algorithm. Moreover, they used association rules trained using UL to provide an initial solution for the search.

The large neighborhood search (LNS) algorithm is a new heuristic algorithm that was first proposed by Shaw \emph{et al.} \cite{ref93} in 1997. Compared with previous heuristics, this algorithm based on the ruin-and-recreate principle \cite{ref94} would explore more complex neighborhoods. The LNS algorithms have outstanding performance in solving various transportation and scheduling problems \cite{ref95}. Therefore, some scholars have considered using the LBO algorithms to improve the LNS algorithms. The learned LNS algorithm for the VRP was first used by Hottung and Tierney \cite{ref96} for the CVRP and split delivery vehicle routing problem (SDVRP) in 2019. They adopted a LBO model as the repair operator for the LNS algorithm. Their model is limited by the number of destroyed fragments; hence, it ineffectively solves large-scale problems. Later, both Chen \emph{et al.} \cite{ref97} and Gao \emph{et al.} \cite{ref98} used a LBO model as a destroy operator of the LNS algorithm to solve the VRP. Chen \emph{et al.} \cite{ref97} used proximate policy optimization (PPO) algorithm to train a hierarchical recursive graph convolution network (GCN) as the destroy operator. Then, they simply inserted nodes removed by the destroy operator into the infeasible solution according to the principle of minimum cost. Experiments on Solomon benchmarks \cite{ref99} and synthetic datasets demonstrated that the model of Chen \emph{et al.} \cite{ref97} performed better than the adaptive LNS (ALNS) algorithm in terms of the solution quality and computational efficiency. Gao \emph{et al.} \cite{ref98} considered the effect of graph topology on the solution, and they used an element-wise graph attention network with edge-embedding (EGATE) as an encoder in which both the node set and arc set contribute to the attention weights. They used a principle similar to that in \cite{ref97} to repair the destroyed solution. Overall, using a LBO model as a destroy operator in the LNS algorithm is more beneficial for generating potential neighbor solutions than as a repair operator.
\subsubsection{Heuristically accelerated LBO algorithms}
Although using the LBO models as assistors can effectively improve the performance of heuristic algorithms, these heuristics still need to elicit knowledge from experts and encode it into the program, which is far from being easy in any applications. Hence, some studies proposed to replace knowledge-modeling in conventional algorithms and used heuristics as assistors of the LBO algorithms to search for solutions. We call these approaches \emph{heuristically accelerated LBO algorithms}.

Reinaldo \emph{et al.} \cite{ref100} proposed a heuristically accelerated distributed deep Q-learning (HADQL) algorithm and compared their model with the ant colony optimization (ACO) algorithm and deep Q network (DQN) on TSP instances. The results indicated that the convergence of their model was fast, but their model required more computation time. Yang \emph{et al.} \cite{ref101} used the LBO algorithm to approximate the dynamic programming function, and their algorithm outperformed other well-known approximation algorithms on large-scale TSPs. Joe and Lau \cite{ref72} used the LBO algorithm to compute the serving cost of each node, then they applied a simulated annealing (SA) algorithm to plan routes to minimize the total cost of the dynamic VRP with time windows and random requirements. Later, Delarue \emph{et al.} \cite{ref102} also used a LBO algorithm to compute the cost of nodes beforehand and their experiments demonstrated that their model achieved an average gap against the well-known solver OR-Tools of 1.7\% on CVRPs. Yao \emph{et al.} \cite{ref73} considered a real-life route planning problem. They regarded the safety of the planned route as one of the objectives and used RL to train the LBO model. They compared their model with EMLS and NSGA-II on a map of New York. Although their framework has good efficiency and optimality, it requires a large amount of time during training.

Chen and Tian \cite{ref103} proposed a LBO model with 2-opt algorithm as the search operator. The model selects a fragment of a solution to be improved according to the region-picking policy and uses the rule-picking policy to select a rewriting rule applicable to the region. Both policies are learnt by the LBO model, and the solution is iteratively rewritten continuously until it converges. Because this framework partially improve the solution, it is significantly affected by the initial solution. Lu \emph{et al.} \cite{ref104} proposed an RL model incorporating several heuristic operators, called the L2I. If the cost reduction after improvement by heuristic operators does not reach the threshold, a random perturbation operator is applied and the operation begins again. Generally, \cite{ref103} and \cite{ref104} are similar because both include improvement and breaking during search process. Although they both obtained good experimental results, \cite{ref104} generated more potential solutions than \cite{ref103} because \cite{ref104} designed a rich set of improvement and perturbation operators. Similar to \cite{ref103}, Costa  \emph{et al.} \cite{refda} and Wu \emph{et al.} \cite{ref106} used an LBO model as a policy network for the 2-opt operator. \cite{refda} built an encoder-decoder framework based on the pointing mechanism \cite{ref107}. Their greatest contribution was that they used different neural networks (NN) to embed node and edge information respectively, and proved that the consideration of edges has an important effect on solving the VRP. The difference in \cite{ref106} from \cite{ref103} is that they did not construct another LBO model to decide the rewritten region, but \cite{ref106} embed the nodes based on the self-attention mechanism before selecting node pairs to apply 2-opt heuristic. Vlastelica \emph{et al.} \cite{ref108} turned solvers into a component of the LBO model to search the optimal solution. They used the LBO model to embed a sequence of nodes into a matrix of pairwise distances to input the solver, then they used the gap between the output of the solver and data label as a loss to optimize the LBO model. Later, Ma  \emph{et al.} \cite{refma} modified the model of Wu \emph{et al.} \cite{ref106} by separating sequence embedding of the current solution from node embedding. Experimental results showed that their LBO model can capture the circularity and symmetry of VRP solutions more effectively. Although above models can obtain better solutions for complex VRPs by making the use of extensiveness of heuristic search, but they need a longer time to search.

In addition to incorporating heuristics as components of the LBO models, scholars have recently proposed to use the LBO algorithm to automatically select or generate heuristics, which can be considered as a type of hyper-heuristic algorithm. Hyper-heuristic algorithms have been used to solve the VRP in many studies \cite{ref109,ref110,ref111}. This kind of algorithms does not search randomly but is guided by other high-level algorithms. Meignan \emph{et al.} \cite{ref112} first used a multi-agent modular model based on the agent metaheuristic framework (AMF) to build a selective hyper-heuristic framework for the CVRP and SDVRP. In their model, each agent builds an AMF model and selects the low-level heuristics. RL and SL mechanisms are used by agents to learn from experience and other agents respectively. Later, Asta and Özcan \cite{ref113} designed a hyper-heuristic algorithm based on apprenticeship learning to produce a new heuristic method for the VRPTW. The same LBO algorithm was used by Tyasnurita \emph{et al.} \cite{ref114} to train the time delay neural network (TDNN) to solve the open vehicle routing problem (OVRP), which does not require vehicles to return to the depot. Moreover, some studies used the LBO algorithms to select heuristics. Both Kerschke \emph{et al.} \cite{ref115} and Zhao \emph{et al.} \cite{ref116}  trained the LBO model as a selector to select the best algorithms for a given TSP instance. Rodr{\'\i}guez \emph{et al.} \cite{ref117} constructed a multi-layer perceptron (MLP) to select the best meta-heuristic for solving the VRPTW, and Martin \emph{et al.} \cite{ref118} used LBO as a selector to select the best parameter settings for randomized Clarke Wright savings for CVRP. Using LBO models as selector of heuristics can leverage complementarity within a set of heuristics to achieve better performance.
\subsection{End-to-end approaches}
End-to-end approaches learn a model to approximate a mapping function between the input (the features of the problem) and the output (the solution of the problem), and the model can directly output a feasible solution on unseen test tasks. The time effeciency for training end-to-end approaches is generally better compared to step-by-step approaches, but they currently encounter challenges in finding high quality solutions and scalability. According to the characteristic (whether the vehicle visits the depot multiple times) of problems which the approach is applied to, end-to-end approaches can be divided into single-path and multi-path planning approaches.
\subsubsection{Single-path planning}
In some VRP variants, the vehicle is not required to return to the depot, such as the TSP, and the LBO algorithms applied to this type of VRPs output a single path that connect all of nodes. Hence, we refer to these LBO algorithms as \emph{single-path planning approaches}.

Standard NNs rely on the assumption of independence among data points and fixed-length inputs and outputs, which is unacceptable for sequence-to-sequence problems \cite{ref120}. States of sequence-to-sequence problems are related in time or space, such as words from sentences and frames from a video. To overcome this shortcoming, recurrent neural networks (RNNs) was designed \cite{ref121}, and other RNN architectures, \emph{long short-term memory} (LSTM) \cite{ref122} and \emph{bidirectional recurrent neural networks} (BRNNs) \cite{ref123} were also introduced for sequence learning in 1997. Later, many studies used these architectures in different sequence-to-sequence problems, such as natural language translation \cite{ref124} and image captioning \cite{ref125}. However, these models compress all the information of the encoder into a fixed-length context vector, which prevents the decoder from focusing on more important information when decoding. Hence, Vinyals \emph{et al.} \cite{ref107}  introduced the attention mechanism \cite{ref126} into the sequence model of \cite{ref124} to enable the decoder to focus on important embeddings and called their model \emph{Pointer Net}. The attention mechanism enables the LBO model to select the contexts that are most closely related to the current state as input. They trained an RNN with a non-parametric softmax layer using SL to predict the sequence of the cities. Although their end-to-end model can directly output the target point of the next step, it is not effective for large-scale TSP. The experimental results also demonstrated that \emph{Pointer Net} produced better solutions when the number of nodes was less than 30, but performed poorly on TSP40 and TSP50.

Vinyals \emph{et al.} \cite{ref107} first proposed an end-to-end model for solving combinatorial optimization problems, but their research needs further improvement. Because SL is undesirable for NP-hard problems, Bello \emph{et al.} \cite{ref127} combined the RL algorithm with \emph{Pointer Net} to solve the TSP in 2016. They proved that their method was superior to that of Vinyals \emph{et al.} \cite{ref107}   on TSP100. Levy and Wolf \cite{ref128} transformed \emph{Pointer Net} to solve other sequence problems in addition to the TSP. They inputed two sequences to two \emph{Pointer Nets} and used a convolution neural network (CNN) to output the alignment scores, which were subsequently converted into distribution vectors using a softmax layer. Later, Deudon \emph{et al.} \cite{ref129} used multiple attention layers and a feedforward layer as encoder to simplify \emph{Pointer Net}. Some studies have also extended the structure of \emph{Pointer Net} to solve other single-path problems. Li \emph{et al.} \cite{ref130} used \emph{Pointer Net} to solve the multi-objective TSP (MOTSP). They first decomposed the multi-objective problem into a series of sub-problems and then used \emph{Pointer Net} to solve each sub-problem. All sub-problems can be solved sequentially by transferring the network weights using the neighborhood-based parameter-transfer strategy. Kaempfer and Wolf \cite{ref131} added a leave-one-out pooling to extend \emph{Pointer Net} to solve the multiple TSP (MTSP), and Le \emph{et al.} \cite{ref132} trained \emph{Pointer Net} using RL to plan a route for cleaning robots. Ma \emph{et al.} \cite{ref133} extended \emph{Pointer Net} to solve the TSP. \cite{ref133} used a graph neural network (GNN) to encode distances of cities as context vectors of the attention mechanism, which is beneficial for solving a large-scale TSP. They also improved their model to a hierarchical framework to solve the TSP with time windows (TSPTW), in which they solved the vanilla TSP in the first level and solved the constraint of time windows in the high level.

Although \emph{Pointer Net} has been widely accepted to solve the TSP since its pioneer contribution, many scholars have proposed other end-to-end models to solve the TSP. As stated in \cite{ref134}, the mentioned neural architectures of \emph{Pointer Net} cannot yet effectively reflect the graph structure of the TSP; therefore, Dai \emph{et al.} \cite{ref134} proposed incorporating a GNN to construct solutions incrementally for the TSP, which is named as \emph{Structure2vec}(S2V) \cite{ref135}, and the embeddings of their model are updated continuously by adding new nodes into currently infeasible solutions. Later, Ottoni \emph{et al.} \cite{ref136} used a set of statistical techniques called the response surface model (RSM) to determine values of the learning rate and discount factor of the LBO algorithm. Although the experiments proved that using RSM can significantly improve the performance of the LBO model, their model need to calculate optimal parameters per instance which limited the generalization of the model. To solve the large-scale TSPs (up to 10000 nodes) in an acceptable runtime, Fu \emph{et al.} \cite{ref137} innovatively introduced a graphic transformation and heat map technique into an end-to-end model. They used graph sampling to abstract sub-graphs from the initial large graph, and the trained LBO model output the corresponding heat map (probability matrix over the edges). Finally, all the heat maps were merged into the final heat map, and the RL algorithm was used to output the optimal solution. The only limitation is that the model only experimented on simulation data but not on other benchmarks or real data. Zhang \emph{et al.} \cite{ref138} proposed an LBO approach for a kind of TSP that allows the vehicle to rejects orders, which was called the TSPTWR. They modified the model of Kool \emph{et al.} \cite{ref74} to output an initial solution for the TSP and then used a greedy algorithm to post-process the solution by rejecting the nodes that violate time windows. Compared with tabu search (TS), this method has a shorter solving time and better results. Because decisions of end-to-end approaches cannot reverse, Xing \emph{et al.} \cite{ref139} combined a GNN and Monte Carlo tree search (MCTS) to solve the TSP. In contrast to previous LBO models that directly made a decision by using a prior probability generated by training, they used MCTS to further improve the decision. Although their model outperforms previous LBO models in test sets of any size, the solving time is much longer.

Groshev \emph{et al.} \cite{ref140} and Joshi \emph{et al.} \cite{ref141} trained a GCN by SL to solve the TSP. What's more, \cite{ref140} further expanded their model by using a trained GCN to guide heuristic algorithms to output solutions and then they used these solutions as labels to retrain the GCN for large-scale TSPs. Prates \emph{et al.} \cite{ref142} also used SL to train an LBO model. They considered the edge weights as features of per instance, and the deviation of their solutions from the optimal can be less than 2\%. Previous LBO models are usually trained and tested on Euclidean space, which makes these models unavailable in instances where nodes are not uniformly distributed. Consequently, Sultana \emph{et al.} \cite{ref143} tested their model in a non-Euclidean space. Although above LBO models trained by SL require fewer samples compared with those LBO models trained by RL, the optimal solutions that are selected as labels have a significant influence on the performance of the models. Joshi \emph{et al.} \cite{ref144} performed controlled experiments between the RL model \cite{ref74} and SL model \cite{ref141} , and they also proved that RL model has better generalization.
\subsubsection{Multi-path planning approaches}Most routing problems are limited by different constraints due to the complex envirnment. In these VRPs, the LBO algorithms need to output multiple path loops since vehicles need return to the depot more than once. We termed these LBO algorithms as \emph{multi-path planning approaches}.

To the best of our knowledge, Nazari \emph{et al.} \cite{ref145} first proposed an end-to-end approach to solve the CVRP. They believed that the order of input is meaningless; thus, they expanded \emph{Pointer Net} by utilizing element-wise projections to map static elements (coordinates of nodes) and dynamic elements (demands of nodes) into a high-dimensional input. The dynamic input is directly fed into the attention mechanism and not complexly computed using an RNN; thus, their model is easier to update embeddings during constructing the solution compared with \emph{Pointer Net}. The optimality of the model was proved by Ibrahim \emph{et al.} \cite{Ibrahim} by comparing it with column generation and OR-Tools on different instances. Later, Kool \emph{et al.} \cite{ref74} introduced a multi-attention mechanism to pass weighted information among different nodes out of consideration for the influence of adjacent structures on solutions. This attention mechanism enables nodes to capture more information from their own neighborhoods; hence, their encoder can learn more useful knowledge to obtain better solutions.

Both Nazari \emph{et al.} \cite{ref145} and Kool \emph{et al.} \cite{ref74} laid the research foundation for the follow-up development of the VRP. Therefore, many scholars either extended their models to solve VRP variants or modified them to obtain better solutions. Peng \emph{et al.} \cite{ref147} considered that node features should be constantly updated during the solving process. Therefore, they built a dynamic attention mechanism model (AM-D) by recoding embeddings at each step. Compared with Kool \emph{et al.} \cite{ref74}, the performance of AM-D is notably improved for VRP20 (2.02\%), VRP50 (2.01\%) and VRP100 (2.55\%). Based the model of Nazari \emph{et al.} \cite{ref145}, Duan \emph{et al.} \cite{Duan} added a classification decoder based on MLP to classify edges. The decoder of Nazari \emph{et al.} \cite{ref145} was used to output a sequence of nodes, while Duan \emph{et al.} \cite{Duan} used this sequence as labels to train another classification decoder for the final solutions. This type of joint training approach outperforms other existing LBO models in solving large-scale CVRP by approximately 5\%. Vera and Abad \cite{ref148} used one more encoder than Kool \emph{et al.} \cite{ref74} to embed vehicle information for the capacitated multi-vehicle routing problem (CMVRP) with a fixed fleet size. The VRPTW has also been widely studied, and the work of Falkner and Schmidt-Thieme \cite{ref149} can be considered the first extension of Kool \emph{et al.} \cite{ref74} to solve the VRPTW. They employed two additional encoders to embed current tours and vehicles for solving large-scale CVRP and produce a comprehensive context for the decoder. Their decoder  would concurrently select the visiting node and the serving vehicle. Their model exhibits strong results on solving VRPTWs, even outperforming OR-Tools.

In the past two years, there has been an increasement in the publication of papers that design end-to-end models for multi-path problems. Similar to Peng \emph{et al.} \cite{ref147}, in 2020, Xin \emph{et al.} \cite{ref150} proposed the dynamic update of embeddings before decoding. However, they considered the computational complexity of the model; they changed the attention weight of the visited nodes in the top layer of the encoder instead of re-embedding. In 2021, Xin \emph{et al.} \cite{ref151} proposed another approach to expand the model of \cite{ref74} by using a multi-decoder with different parameters to train multiple policies and select the best one to decode at each step. Experiments indicated that these innovations are useful in improving the original LBO model. Zhang \emph{et al.} \cite{ref152} used the model of Kool \emph{et al.} \cite{ref74} to solve multi-vehicle routing problems with soft time windows (MVRPSTW). They considered vehicles as multi-agents, where all agents share one encoder but use different decoders. Zhao \emph{et al.} \cite{ref153} modified the model proposed by Nazari \emph{et al.} \cite{ref145} by using a routing simulator to generate data and update both the dynamic state and mask for the CVRP and the VRPTW. Furthermore, they applied a local search to improve solutions of the LBO model, and they demonstrated that combining the LBO algorithms and heuristic search can be a general method of solving combinatorial problems. In contrast to previous LBO models, which are based on the standard policy gradient method, Sultana \emph{et al.} \cite{sultana} first proposed to add an entropy regularization term to encourage the exploration for solving VRPs, thereby avoiding the limitation that the LBO model can easily converge too rapidly to a poor solution. They experimentally demonstrated that the policy with the highest entropy was easier to find a satisfactory solution. Drori \emph{et al.} \cite{ref154} transformed the structure of the VRP into a line graph by defining each edge in the primal graph corresponding to a node in the line graph, and computed edge weights as node features. To effectively solve dynamic and stochastic VRP (DS-VRP), Bono \emph{et al.} \cite{ref155} proposed a multi-agent model by adding two attention networks based on \cite{ref74} to encode vehicle’s features and last decision. Their model presented favorable results for small-scale DS-CVRP and DS-CVRPTW compared with OR-Tools. However, the model exhibited poor generalization when testing on deterministic CVRPs and CVRPTWs. Lin \emph{et al.} \cite{ref156} incorporated the model of Nazari \emph{et al.} \cite{ref145} with a graph embedding layer to solve the electric vehicle routing problem with time windows (EVRPTW). They trained the model using the modified REINFORCE algorithm proposed by Kool \emph{et al.} \cite{ref74}, but the maximum number of nodes in the test instance was only up to 100. Recently, Li  \emph{et al.} \cite{ref157}  modified the model proposed by Kool \emph{et al.} \cite{ref74}  to solve the pickup and delivery problem (PDP), which has priority constraints and specific pairing relations. To learn complex relations and precedence among nodes of different roles, they added another six types of attention mechanisms based on the original attention mechanism. Later, aiming at that most end-to-end approaches are designed for homogeneous vehicle fleet, Li \emph{et al.} \cite{refli} proposed to add a vehicle decoder to minimize the travel time among all heterogeneous vehicles based on the model of Kool \emph{et al.} \cite{ref74}. Their LBO model would select both a serving vehicle and a visiting node rather than solely selecting the next node to visit at each step.

Many scholars have proposed other novel end-to-end frameworks for VRPs with different objectives. Li \emph{et al.} \cite{ref158} and Yu \emph{et al.} \cite{ref159} proposed a model to plan online vehicle routes to minimize computation time. The differences between two papers lie in the architecture and problem background. Li \emph{et al.} \cite{ref158} first used an LSTM network to predict future traffic conditions and then used a double-reward value iterative network to make decisions. However, Yu \emph{et al.} \cite{ref159} planned routes by improving \emph{Pointer Net} \cite{ref107}. In addition, Li \emph{et al.} \cite{ref158} trained their model on data of 400,000 taxi trajectories in Beijing, whereas Yu \emph{et al.} \cite{ref159} applied to the green logistics system and trained their model on the traffic data of Cologne, Germany. Balaji \emph{et al.} \cite{ref160} combined distributed prioritized experience reply \cite{ref161} with DQN to maximize the total reward of the VRP. Although DQN is widely used in LBO models, it requires a significant amount of time to converge because the information update lag of the experience replay. With respect to this limitation, Mukhutdinov \emph{et al.} \cite{ref162} proposed using SL to generate preliminary Q values for nodes and they applied this approach to minimize the cost of the packet routing problem. RamachandranPillai \emph{et al.} \cite{ref163} designed an adaptive extended spiking neural P system with potentials (ATSNPS) to determine the shortest solutions for the VRPTW. They experimented on supermarket chain instances and demonstrated that their model can obtain better solutions. Sheng \emph{et al.} \cite{ref164} according to the principle of maximizing the total benefit, introduced global attention \cite{ref165} to modify \emph{Pointer Net} \cite{ref107} to solve the VRP with task priority and limited resources (VRPTPLR). Cao \emph{et al.} \cite{ref166} first used the RL model to solve the stochastic shortest path (SSP) problem requiring on-time arrival. They used the Q-value to represent the probability of arriving on time and set the discount factor of reward as 1 to maximize the probability of arriving on time. Chen \emph{et al.} \cite{ref167} built an LBO model for an autonomous vehicle fleet and post-processed routes for minimizing the energy cost of the entire fleet. However, the environment of their model was too idealistic, and they did not consider the dynamics in the real world.

In addition to designing models for the VRP with a single objective, some studies aimed at achieving multiple objectives simultaneously, which is difficult for traditional algorithms. To minimize driving time and route length, Kalakanti \emph{et al.} \cite{ref168} proposed a framework with two stages, which included clustering by heuristics and route planning by Q-learning. However, experiments on three VRP variants demonstrated that their model performed poorly in a stochastic setting. To minimize the tour length and cost of the ride-sharing field, Holler \emph{et al.} \cite{ref169} used a MLP to compute the pooling weights and selected action based on the pooling mechanism. They used the DQN and PPO algorithm to train the LBO model respectively, and experiments demonstrated that DQN is more efficient.
\subsection{Analysis of literatures}
\begin{figure}[htb]
	\begin{center}
		\includegraphics[width=3.2in,height=2in]{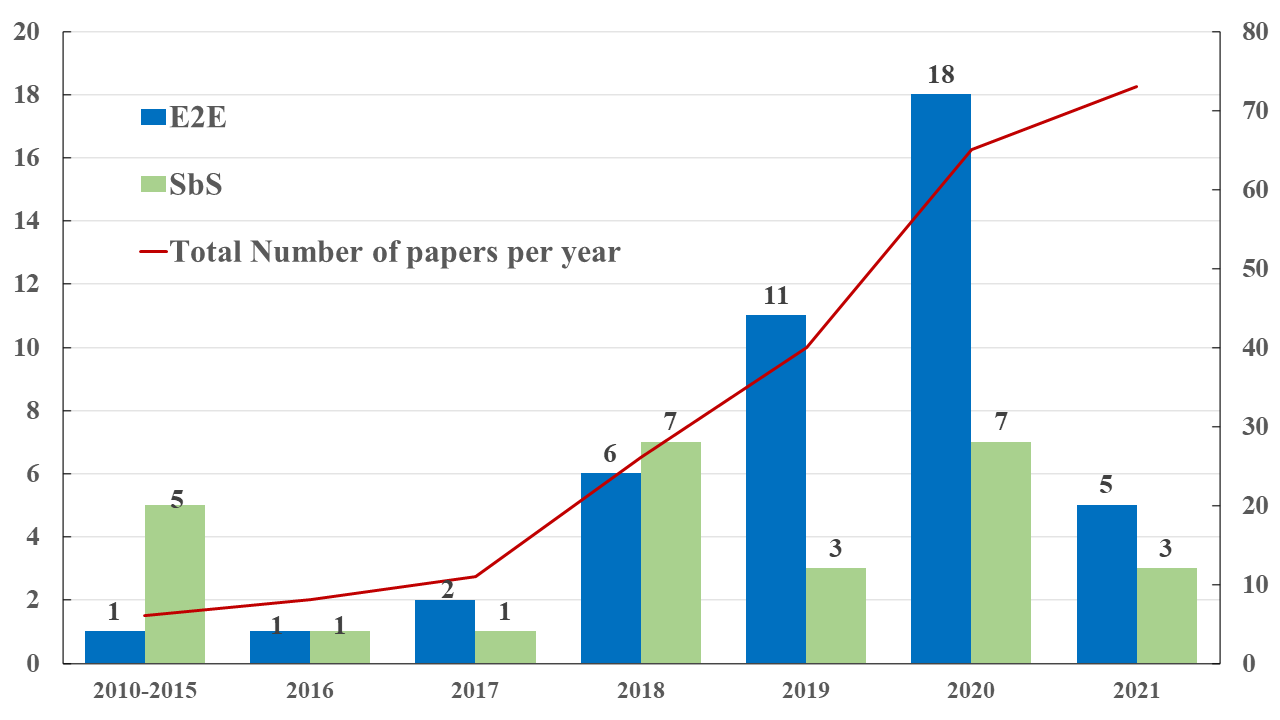}
	\end{center}
	\caption{Distribution of published papers per year for the VRP (the deadline for statistical data in 2021 is September).}\label{Fig:4}
\end{figure}

As we indicated previously, many papers that using the LBO algorithms to solve different VRP variants have been published (Fig.~\ref{Fig:4}). There are 14 and 25 relevant papers were respectively published in 2019 and 2020. As we can observe, there has been a rapid expansion in this field over the last two years. This is primarily because of the following reasons: 1) an increasing number of scholars have proved that the LBO algorithm is competitive in solving combinatorial optimization problems; 2) with the development of economic globalization, transportation efficiency has become a key factor affecting company profits; 3) recent VRP studies have been characterized by large-scale online planning and complex constraints. These factors significantly promote the LBO algorithms in solving the VRP. Another aspect worth noting in Fig.~\ref{Fig:4} is that scholars tended to use step-by-step approaches in the initial research phase of the field. However, they preferred end-to-end approaches after 2017, which is primarily owing to the history-making success achieved by Alpha GO. However, with the recent in-depth research in using the LBO algorithms for the VRP, many literatures have demonstrated that combinning LBO model with other algorithms is more effective for complex optimization problems. Hence, the research on the step-by-step methodes has been revived.

\begin{figure}[H]
	\includegraphics[width=3.4in,height=2in]{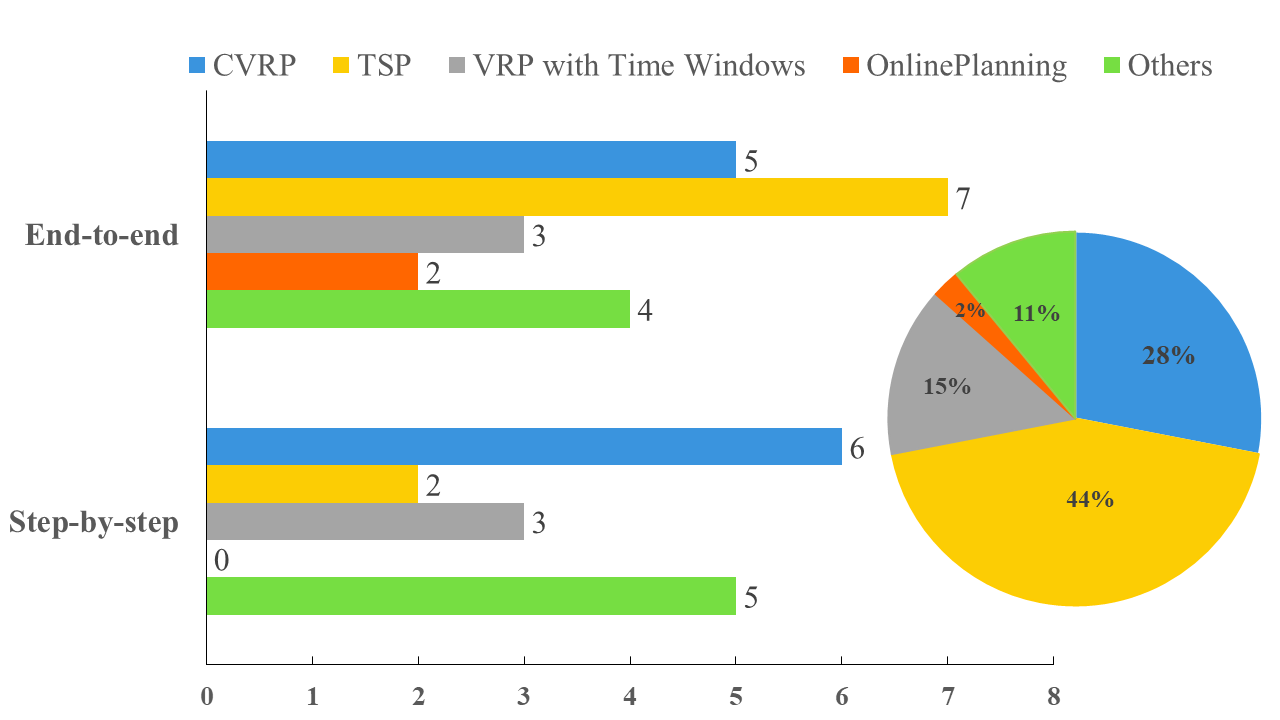}
	\caption{Distribution of different variants.}\label{Fig:7}
\end{figure}

\begin{figure}[H]
	\includegraphics[width=3.5in]{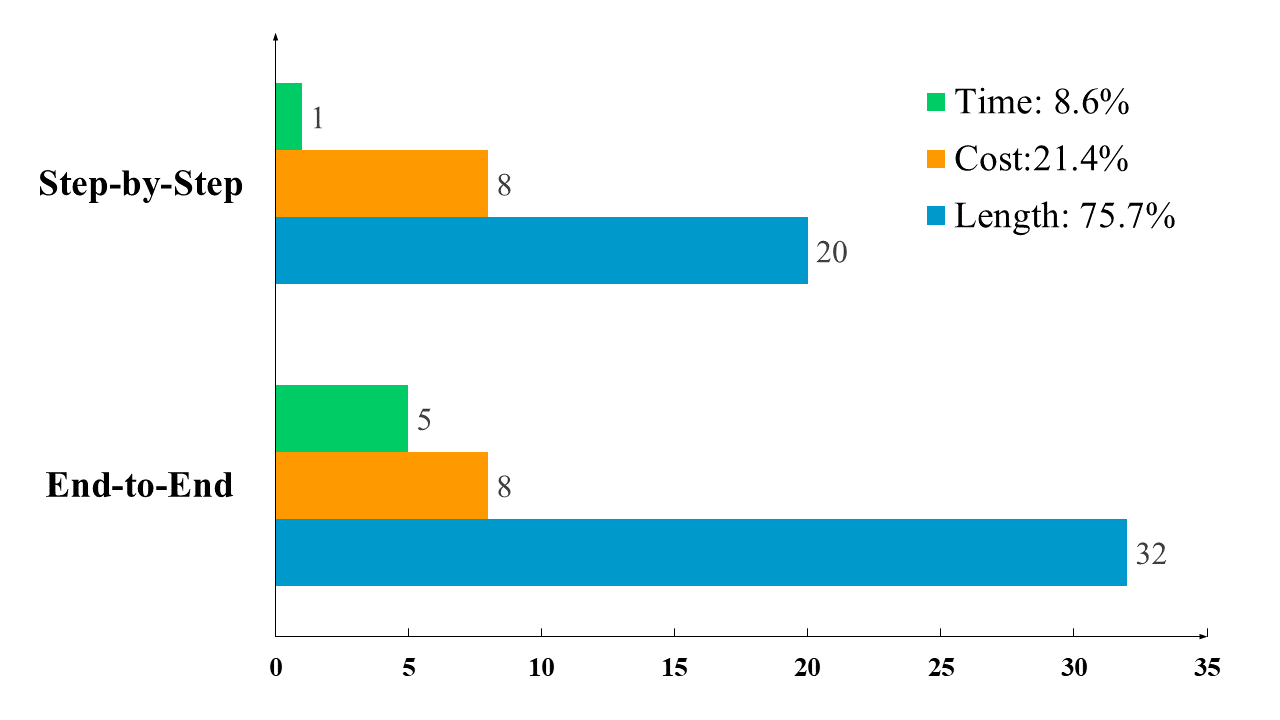}
	\caption{Distribution of objective functions.}\label{Fig:6}
\end{figure}

As shown in the pie chart on the right of Fig.~\ref{Fig:7} , most of studies focus on the TSP (approximately 44\%). As a widely studied variants of the VRP, the TSP is a basic graph problem; therefore, scholars tend to test a new model on the TSP and then extend the model to other VRP variants. The second widely studied variant is the CVRP (approximately 28\%), which has a simple mathematical model but strong flexibility. Both the CVRP and TSP primarily target minimizing the tour length; hence, length is the most studied objective among the distribution in the objectives of the literatures (approximately 75.7\%, Fig.~\ref{Fig:6}). Note that the sum of percentages in Fig.~\ref{Fig:6} is greater than 1 because some studies are multi-objective.
\vspace{-0.3cm}
\vspace{-0.5cm}

We summarize the characteristics of different VRP problems by synthesizing and analyzing the referenced papers. The four evaluation indicators are described as follows.
\begin{itemize}
\item \textbf{High-complexity}. High-complexity refers to VRPs with multiple constraints or objectives, and we quantifiably represent the practicability of an LBO model by complexity of its application problem because the VRP in the real world often has multiple constraints or objectives. In this paper, we consider that problems with two or more objectives or constraints are high complex. For example, the TSPTW with rejection solved by Zhang \emph{et al.} \cite{ref138} needs to minimize the tour length and the rejection rate; the VRP solved by Bono \emph{et al.} \cite{ref155} requires time windows and stochastic demands to be satisfied simultaneously. We considered that the above problems are complex, and we compared the number of studies of two class approaches applied to complex problems. We observed that step-by-step approaches are more likely to be selected for problems with high-complexity.
\item \textbf{Stochastic}.  We define the VRPs whose customer's demands, time windows or other uncertain elements follow some probability distribution models as stochastic problems. For example, Balaji \emph{et al.} \cite{ref160} considered a VRP variant of on-demand delivery, whose orders were generated with a constant probability; Joe and Lau \cite{ref72} considered a real-time VRP in urban logistics in which customers and orders were randomly added or cancelled. Step-by-step approaches have limitations in solving stochastic VRPs since this type of the LBO algorithms needs customers and travel costs to be known in advance and longer solving time. Compared with step-by-step approaches, end-to-end approaches have the advantages of flexibility and time efficiency under stochastic environments, which can quickly responce to changes by utilizing knowledge extracted from past training experience.
\item \textbf{Timeliness}. Timeliness refers to those VRPs that require to plan a tour with the least time, which is a significant characteristic of real-world VRPs. We consider problems that use LBO approaches to minimize tour time are timeliness, such as in \cite{ref159}. Overall, end-to-end approaches are easily used for problems with timeliness because of their quick solution speed.
\item \textbf{Fuzzy}. With the development of the VRP, many new VRP variants have been proposed. We use fuzzy to define these new problems because researchers do not have much expertise in these problems. The more likely a type of the LBO approaches is used to solve new problems indicates that it has a more generic modeling framework. There are more fuzzy problems are solved by end-to-end approaches among our referred literatures since the learning process of these approaches do not require abundant domain knowledge.\end{itemize}

Generally, end-to-end approaches are suitable for VRPs with stochastic or time requirements; step-by-step approaches can solve more complex problems effectively.

\begin{figure*}[htbp]
	\centering
	\includegraphics[width=5in,height=5.7in]{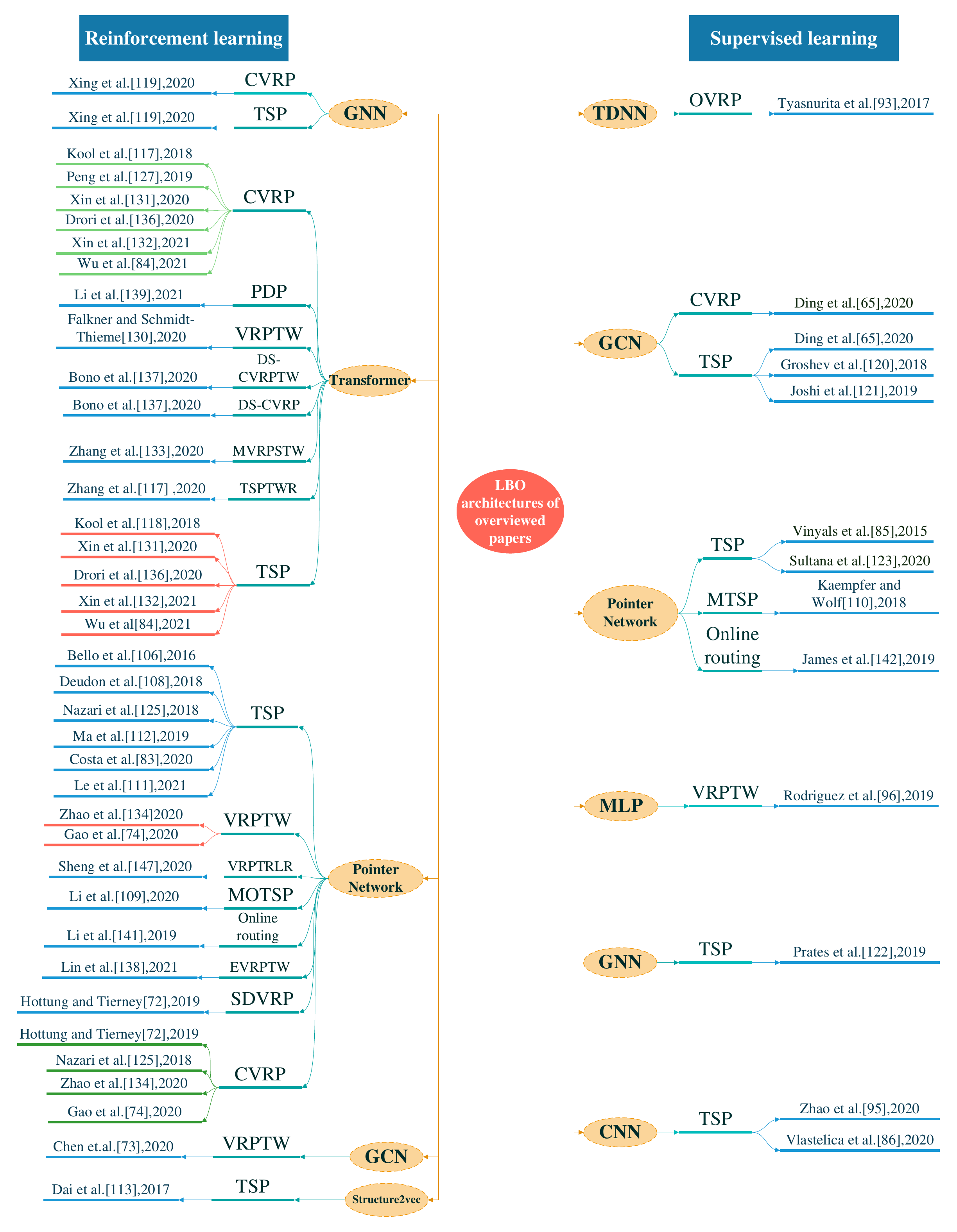}	
	\caption{Overview of LBO architecture of overviewed papers (\cite{ref170}).}\label{Fig:12}
\end{figure*}

In addition to analyzing the applied problems of the LBO algorithms, it is interesting to compare their models. The previous review clearly indicates that many studies used the encoder-decoder framework; therefore, we compared these encoder-decoder models (see Table~\ref{tab:1}, where MHA refers to the multi-head attention layers and FF refers to the feedforward network). We observed that the encoder-decoder framework is more commonly used in end-to-end approaches, and most differences of frameworks are in the encoder. This is probably because if scholars require different information, they must use different NNs to extract related features from the input. If authors seek to incorporate more features in addition to the node coordinates and demands, they often consider the GCN as a good option. Conversely, there are two main types of decoders: RNN with a pointing mechanism (PM) (Vinyals \emph{et al.} \cite{ref107}) and the other is composed of several multi-head attention sublayers (MHA) (Kool \emph{et al.} \cite{ref74}). In terms of learning manner, Vinyals \emph{et al.} \cite{ref107} used SL to train the model, whereas all the others used RL. This is expected because the VRP is an NP-hard problem, and it is difficult to obtain labels for training data. We also present an overview of the LBO architectures in Fig.~\ref{Fig:12}. We first classify NN models according to training algorithms. Then, we list solving problems of different structures together with the corresponding literature and their publication time.

To clearly compare the literature, we list the main content of the referenced papers, including model features, baselines, and benchmarks in experiments at the end of the paper (see Table~\ref{tab:2}). For convenience, we abbreviate end-to-end approaches as E2E and step-by-step approaches as SbS in this table. Benchmarks can frequently be divided into simulation data and data from the literature or the real world; we refer to the latter collectively as real data.
\vspace{-0.5cm}

\begin{table*}[htbp]
	\centering\vspace{0.5cm}
	\caption{Comparison of encoder-decoder frameworks among referred models}
		\resizebox{\textwidth}{!}{
			\begin{tabular}{cccccm{12.325em}<{\centering}c}
				\toprule[2pt]
				Approaches & Literature & Problem & Encoder & Embedding features & \multicolumn{1}{c}{Decoder} & {Learning manner} \\
				\midrule
				\multicolumn{1}{c}{\multirow{24}[1]{*}{End-to-end Approaches}} & 
				 \cite{Duan}  & CVRP  & GCN   & nodes \&distance matrix & RNN + PM & RL+SL \\
				& \cite{ref74}  & CVRP  & MHA   & nodes & MHA   & RL \\
				& \cite{ref107} & TSP   & RNN   & nodes & RNN + PM & SL \\
				& \cite{ref127} & TSP   & RNN   & nodes & RNN + PM & RL \\
				& \cite{ref128} & TSP   & CNN   & nodes & RNN   & SL \\
				& \cite{ref130} & TSP   & RNN   & nodes & RNN + PM & RL \\
				& \cite{ref132} & TSP   & GNN   & nodes & RNN + PM & RL \\
				& \cite{ref133} & TSP   & GNN   & nodes & RNN + PM & RL \\
				& \cite{ref138} & TSPTWR & MHA   & nodes & MHA   & RL \\
				& \cite{ref143} & TSP   & CNN   & nodes & RNN + PM & SL \\
				& \cite{ref145} & CVRP  & GCN   & nodes & RNN + PM & RL \\
				& \cite{ref147} & CVRP  & MHA   & nodes & MHA   & RL \\
				& \cite{ref148} & CMVRP & GCN   & nodes & RNN + PM & RL \\
				& \cite{ref149} & CVRPTW & MHA   & nodes \& tour \& vehicles & MHA   & RL \\
				& \cite{ref150} & CVRP,TSP & MHA   & nodes & MHA   & RL \\
				& \cite{ref152} & MVRPSTW & MHA   & nodes & MHA   & RL \\
				& \cite{ref153} & CVRP,VRPTW & GCN   & nodes & RNN + PM & RL \\
				& \cite{ref154} & CVRP,TSP & MHA   & edges & MHA   & RL \\
				& \cite{ref155} & DS-CVRP,DS-CVRPTW & MHA   & nodes \& vehicles & MHA   & RL \\
				& \cite{ref156} & EVRPTW & GCN   & nodes \& battery \& time \& vehicles & RNN + PM & RL \\
				& \cite{ref157} & PDP   & MHA   & nodes & MHA   & RL \\
				& \cite{ref159} & Online VRP & GNN   & nodes & RNN + PM & RL \\
				& \cite{ref164} & VRPTPLR & RNN   & nodes & RNN + PM & RL \\
				& \cite{ref156} & EVRPTW & GCN   & nodes & RNN + PM & RL \\
				\midrule
				\multicolumn{1}{c}{\multirow{3}[2]{*}{Step-by-step Approaches}} 
				& \cite{ref96}  & CVRP, SDVRP & Two linear layers & tour  & FF    & RL \\
				& \cite{ref98}  & CVRP, CVRPTW & EGATE & nodes \& edges & RNN + PM & RL \\
				& \cite{refda} & TSP   & GCN   & tour  & PM +Max-pooling \& FF +Mean-pooling & RL \\
			\toprule[1pt]
			\end{tabular}%
		}
		\label{tab:1}%
	\end{table*}%
~\
\section{Experimental study}\label{sec5}
\subsection{Experimental Setting}
\textbf{Experimental algorithms.} As described in Section~\ref{sec3}, the research of Lu \emph{et al.} \cite{ref104} (L2I) and Chen and Tian \cite{ref103} (Rewriter) are the recent research trends of the step-by-step approaches, and both papers are also among the most cited step-by-step approaches. Kool \emph{et al.} \cite{ref74} (AM) made a significant contribution to solving the VRP by using an end-to-end framework. Many studies have modified the model of Kool \emph{et al.} \cite{ref74}, and the model of Xin \emph{et al.} \cite{ref150} (ASWTAM), proposed in 2020, is one of the representatives. ASWTAM designs a step-wise update embedding mechanism, which is beneficial for the original AM model to focus on useful inputs and determine the optimal solution. To analyze the characteristics and limitations of different LBO models, we test these four representative LBO algorithms on different sizes of instances, and we compare them with other algorithms in this section. We select three classical meta-heuristics (ACO, TS and LNS), and two well-known solvers (Gurobi and OR-Tools) as baseline algorithms. The best parameters of different scales of the problems are determined through multiple experiments. To limit the total time expended when solving the problem, we modify the time limit parameter of Gurobi to 1800 seconds.

\textbf{Problem Details.} We evaluate four LBO models and baseline algorithms on the CVRP of different scales, and we do not distinguish the CVRP and VRP in this section.

\textbf{Data generation.}

(1)\textbf{Training set.} We consider three training instances, Euclidean VRP with 20, 50, and 100 nodes, named VRP20, VRP50, and VRP100, respectively. For all tasks, the location of each customer and the depot are uniformly sampled in the unit square [0,1]$^2$, and the demand of each customer is also uniformly sampled from the discrete set \{1, 2, …, 9\}. The capacities of a vehicle are 30, 40, and 50 for N = 20, 50, and 100, respectively.

(2)\textbf{Test set.} We test different algorithms using three types of test data: 1) \textbf{Set 1}, following the same rules as the training set, we newly generate 300 instances for the three-scale VRP. 2) \textbf{Set 2} also contains 300 VRP instances with \emph{n}=20, 50, and 100, but the locations of the nodes are sampled from the gamma distribution ($\alpha$=1, $\beta$=0). 3) \textbf{Set 3} contains nine benchmarks from Uchoa \emph{et al.} \cite{ref171}, whose nodes ranged from 100 to 200.

\textbf{Hyperparameters setting.} The hyperparameters of the selected LBO models are the same as those in literatures to ensure the validity of the experimental results as much as possible. We set the random seed at 1234 to ensure the consistency of training data, and we set the batch size during testing to 1 to better compare the running time. Table~\ref{tab:3} lists the other hyperparameters of the training model. We conduct all the experiments on Python software on a computer with a Core i7-9800x 3.8-GHz CPU, 16 GB memory, Windows 10 operation system, and a single 2080Ti GPU.
\begin{table}[htbp]
	\centering
	\caption{The hyperparameters of LBO models}
	\resizebox{\columnwidth}{!}{
	\begin{tabular}{m{9.69em}<{\centering}cccc}
		\toprule[2pt]
		Parameters & \multicolumn{1}{m{4.19em}<{\centering}}{AM} & \multicolumn{1}{m{4.19em}<{\centering}}{Rewriter} & \multicolumn{1}{m{4.19em}<{\centering}}{L2I } & \multicolumn{1}{m{4.19em}<{\centering}}{ASWTAM} \\
		\midrule
		The number of epochs & 100   & 10    & 40000 & 100 \\
		The size of training set & 1280000 & 100000 & 2000  & 1280000 \\
		Batch size & 512   & 128   & 1000 & 512 \\
		Learning rate & 0.001 & 5.00E-05 & 0.001 & 1.00E-04 \\
		\bottomrule[2pt]
	\end{tabular}%
	\label{tab:3}%
}
\end{table}%

\textbf{Evaluation metrics.} To better evaluate LBO models, we use multiple evaluation metrics that are widely used in the RL research community \cite{ref40,ref160}. In addition, since each type of VRP in set 1 and 2 only contains 300 instances, we use the Wilcoxon signed-rank test and Friedman test on test results to infer the holistic performance of different LBO models.\begin{itemize}
\item Training time and occupy memory.
\item Length of solutions, optimal gap, and solving time in per-instance.
\item Rank obtained by Friedman test and \emph{p}-value obtained from the Wilcoxon test.
\end{itemize}

\textbf{Experimental design.}

To fully evaluate the effect of LBO models, we compare the experimental algorithms from three aspects: time efficiency, scalability, and optimality. The comparison experiments can be divided into three parts:
\begin{description}
	\item[\textbf{Part \Rmnum{1}}]To validate the learning effectiveness of the LBO models, we train the LBO models on the training set and compare the training time, occupied memory, and learning curves among the LBO models.
	\item[\textbf{Part \Rmnum{2}}]To validate the time efficiency and optimality of the LBO models, we test algorithms on set 1 and 2 and compare the solution length, solving time, and standard deviation. In addition, we use the Wilcoxon signed-rank test and Friedman test to analyze statistical results in depth.
	\item[\textbf{Part \Rmnum{3}}]To validate the scalability of LBO models to larger problems, we test LBO models trained on VRP20 on set 3. We use the solution length, solving time, and optimal gap as evaluation indicators.
\end{description}
\subsection{Results and Analysis}
\subsubsection{Comparison and discussion of Part \Rmnum{1}}
\begin{figure}[htb]
	\begin{center}
		\includegraphics[width=3.5in]{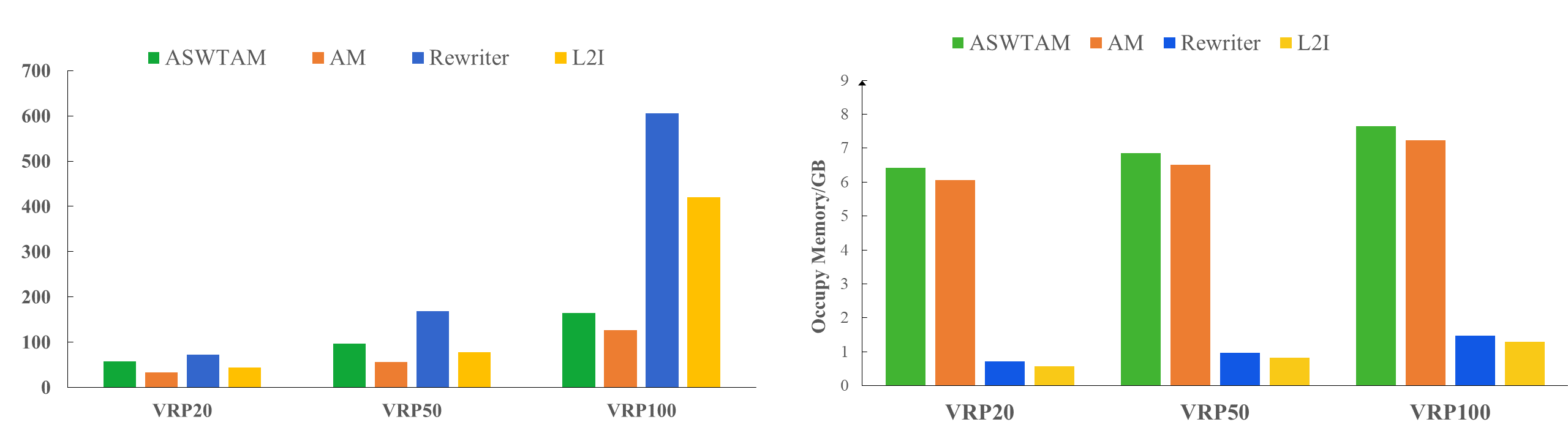}
	\end{center}
	\caption{Training time and occupy memory of LBO models on VRP20, VRP50, and VRP100.}\label{Fig:8}
\end{figure}

We compared the training time (in hours) of the four LBO models on VRP20, VRP50, and VRP100. The left bar chart of Fig.~\ref{Fig:8} shows that end-to-end approaches frequently requires less training time than step-by-step approaches. AM requires the least training time, whereas Rewriter requires the most training time. This is primarily because these step-by-step approaches require to be combined with heuristic search to determine optimal solutions; therefore, they require more time to constantly experience the circulation of generation, evaluation, and evolution during training. While end-to-end approaches use NNs to replace this complex circulation, they can cost less time to learn. Comparing Rewriter with L2I, although Rewriter uses only 2-opt to generate neighborhood solutions and L2I uses six heuristics, the training size of Rewriter is approximately 50 times than that of L2I. Therefore, Rewriter requires more time than L2I. Comparing two end-to-end approaches, ASWTAM costs longer training time since it needs to step-wise update embeddings at each step.

We also recorded the occupied memory of the LBO models during training (right bar chart of Fig.~\ref{Fig:8}), and we observed that step-by-step approaches require less memory than end-to-end approaches; e.g., L2I requires only approximately 1.2892 GB on VRP100 and ASWTAM requires 9.07 GB. This result reveals that end-to-end approaches can fully utilize advanced computing hardware to reduce training time but require the computer to have a large amount of memory to store sizable data generated by parallel computing. In addition, we observed that LBO models require more memory as the size of the problem increases. This is apparent because a large-scale VRP has high input dimensions, and solving it requires deeper NNs with more parameters. However, the problem size remains a challenge for the LBO algorithms to settle because of the curse of dimensionality and the limitations in computational resources. Comparing L2I with Rewriter, Rewriter requires more memory during training because it uses another policy network to select segments to be rewritten in addition to a rule-defining policy network. Similarly, we observed that ASWTAM requires more memory than AM. This is because ASWTAM requires step-wise update embeddings but the embeddings of AM are fixed.

\begin{figure}[htb]
	\begin{center}
		\includegraphics[width=3.5in]{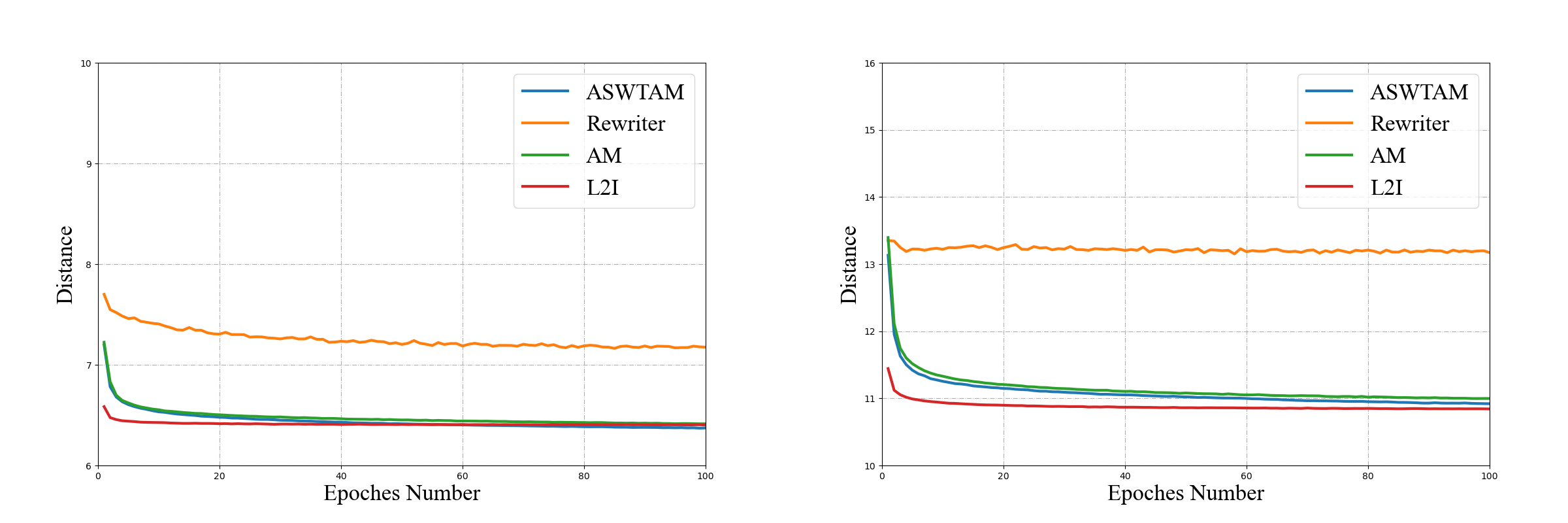}
	\end{center}\vspace{-3mm}
	\caption{Learning curves of LBO models on VRP20 and VRP50.}\label{Fig:9}
\end{figure}

We depicted the comparison of learning curves in Fig.~\ref{Fig:9}, and we defined the average solution length of samples of a batch as distance. The learning curve is an important index for evaluating the learning ability of an LBO model, and it can provide much information about the LBO model. The earlier the turning point of the curve tends to be stable, the fewer samples are required for model training. In addition, the distance in the training process predicts the optimization of the model on the training data, whereas a lower value indicates a better performance of the LBO model. Comparing the two approaches, we can conclude that step-by-step approaches can converge faster than end-to-end approaches. Among them, L2I has the best learning performance because its learning curve reaches the turning point earlier, which proves that using heuristics as search operators can effectively improve the learning effectiveness of LBO models. This also explains why step-by-step approaches require less training data than end-to-end approaches. In addition, we can speculate that there is a mutual promotion between LBO models and heuristic algorithms, but exceedingly few heuristics may reduce the learning ability of LBO models, similar to Rewriter. Additionally, note that the learning curve of ASWTAM is below the AM as the training progressed. This proves that the input features have an important influence on the learning ability of the model because ASWTAM modifies AM by dynamically updating the embeddings.
\vspace{-0.5cm}

~\
\subsubsection{Comparison and discussion of Part \Rmnum{2}}Table~\ref{tab:4} and \ref{tab:5} show the optimization of the LBO algorithms on test sets 1 and 2, respectively. For the columns in Table~\ref{tab:4} and \ref{tab:5} , column 1 shows the algorithms, and columns 2-4 respectively list the average tour length (mean), standard deviation (std), and average solving time (time) used by each algorithm for instances with {$n$}=20. Columns 5-7 and 8-10 respectively list the same information of experimental algorithms on the instances with {$n$} = 50 and 100. Note that we tested two types of search policy for AM for a comprehensive comparison.

First, the tables show that L2I has the minimum mean on testing instances except for VRP20, and its results are even better than those of OR-Tools. Furthermore, we observed that the standard deviations of L2I and Rewriter are always smaller than those of OR-Tools, LNS and ACO These results indicate that incorporating the LBO algorithms within the heuristic search can result in a stable performance and strong generalization of the models. Several studies \cite{ref129,ref143,ref153} have indicated this positive effect. Second, we observed that TS and LNS perform below ASWTAM and AM (greedy) in VRP50 and VRP100 on set 1, and ASWTAM is better than OR-Tools in VRP100. However, end-to-end approaches have disappointed performance on set 2, although their solutions are still better than that of ACO. Therefore, we can conclude that end-to-end approaches have a strong dependence on the distribution of data. In addition, comparing L2I to end-to-end approaches, although L2I outperformes ASWTAM and AM in terms of the quality of the solution, end-to-end approaches have an advantage in computational time, e.g., AM (greedy) requires only \textbf{0.93 s} and ASWTAM requires \textbf{1.82 s} on VRP100 from set 1, but L2I requires 25.25 s. Third, we observed that different search methods are applicable to different types of test data. Comparing Table~\ref{tab:4} and \ref{tab:5}, we can conclude that the sampling search method outperforms the greedy search method on data obeying the gamma distribution, but the greedy search method can obtain better solutions for data obeying the same distribution as the training data. This is because the greedy search selects the best action at each step according to the training experience, whereas the sampling search selects the best from many sampled solutions \cite{ref133}. Hence, a greedy search is more dependent on the data distribution. Finally, comparing L2I with Rewriter, we observed that L2I has better optimization than Rewriter. Thus, more search operators are beneficial to searching for a larger solution space, and the optimal solution is more likely to be determined. However, note that an excessive number of heuristic operators result in the solving time of L2I being twice as that of Rewriter. We also compared two LBO models of end-to-end approaches and concluded that ASWTAM is better than AM in quality of solutions. ASWTAM adopts a dynamic embedding mechanism, and dynamic embedding aids the network in capturing the real-time characteristics of the environment.

\begin{figure}[htb]
	\begin{center}
		\includegraphics[width=3.6in]{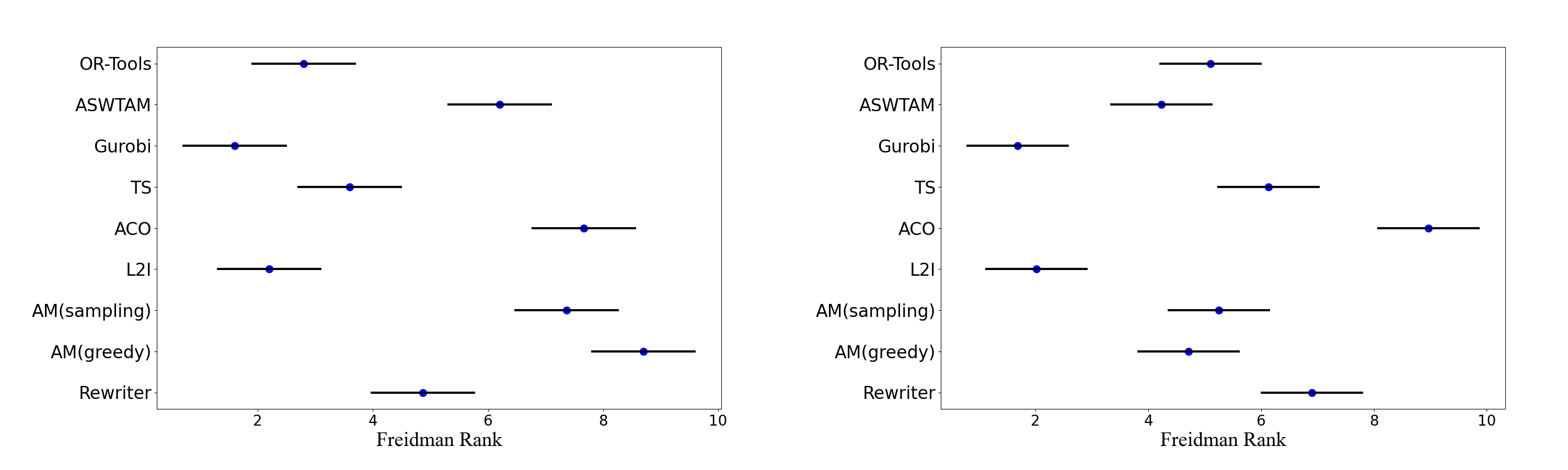}
	\end{center}
	\caption{Friedman test on VRP20 from different datasets (left from set 1, right from set 2).}\label{Fig:10}
\end{figure}

To further illustrate the overall performance of the experimental algorithms across all test cases, we performed nonparametric tests on VRP20 from set 1 and 2, and the results are shown in Fig.~\ref{Fig:10} and Table~\ref{tab:6}. Fig.~\ref{Fig:10} shows the results of the Freidman test and uses Freidman Rank as the abscissa. The smaller the rank value is, the better the algorithm performs in all instances. Table~\ref{tab:6} shows the results of the Wilcoxon test and column 1 represents the tested algorithms. The Wilcoxon test first computes the difference between the two algorithms on a set of instances and then obtains the corresponding rank by sorting the absolute values of the differences. Column 2 of Table~\ref{tab:6} summarizes the rank of the positive differences, and column 3 summarizes the rank of negative differences. The gap between the two sums represents the difference between the two algorithms, and the \emph{p}-value in the final column is used to evaluate this difference quantitatively. Under the confidence degree $\alpha$=0.05, a \emph{p}-value greater than 0.05 means that there is no significant difference between the two algorithms.

From the Friedman test (in Fig.~\ref{Fig:10}) on VRP20, we observed that L2I is second only to Gurobi in VRP20, and the results of the Wilcoxon test of the two algorithms in Table~\ref{tab:6} indicates that L2I is not significantly different from Gurobi in VRP20 at a confidence coefficient of 0.05. Moreover, we observed that the performance of ASWTAM is not significantly different from AM on set 1, but AM is distinctly inferior to ASWTAM on dataset 2. This proves that AM is effectively improved by ASWTAM. It is worthy noting that the \emph{p}-value of OR-Tools and L2I is 1 in set 2, although this value is 0.000002 in set 1. Similar results are observed in the comparison of the other three LBO models with OR-Tools, in which LBO models achieve \emph{p}-values $>$0.05 in set 1 but $<$0.05 in set 2. This indicates that LBO models have limitations when applied to the problem whose data distribution is different from the training set.
\vspace{-0.5cm}
\vspace{-0.5cm}
\begin{table}[htbp]
	\centering%
	\caption{Wilcoxon test results among different algorithms for $\alpha$=0.05 on VRP20 from different date set.}
	\resizebox{\columnwidth}{!}{
		\begin{tabular}{m{12em}<{\centering}cccccccc}
			\toprule[2pt]
			\multirow{2}[4]{*}{\textbf{Comparing Algorithms}} &       & \multicolumn{3}{m{12.945em}<{\centering}}{\textbf{Data Set 1}} &       & \multicolumn{3}{m{12.63em}<{\centering}}{\textbf{Data Set 2}} \\
			\cmidrule{3-5}\cmidrule{7-9}    \multicolumn{1}{c}{} &       & \multicolumn{1}{m{4.19em}<{\centering}}{R+} & \multicolumn{1}{m{4.19em}<{\centering}}{R-} & \multicolumn{1}{m{4.565em}<{\centering}}{\textit{p-value}} &       & \multicolumn{1}{m{4.19em}<{\centering}}{R+} & \multicolumn{1}{m{4.19em}<{\centering}}{R-} & \multicolumn{1}{m{4.25em}<{\centering}}{\textit{p-value}} \\
			\midrule
			L2I vs OR-Tools &       & 465   & 0     & 0.000002  &       & 267   & 198   & 0.471592  \\
			L2I vs Gurobi &       & 168   & 267   & 1.000000  &       & 82.5  & 382.5 & 1.000000  \\
			L2I vs TS &       & 465   & 0     & 0.000002  &       & 465   & 0     & 0.000002  \\
			Rewriter vs ACO &       & 465   & 0     & 0.000002  &       & 465   & 0     & 0.000002  \\
			Rewriter vs TS &       & 109   & 356   & 1.000000  &       & 3     & 462   & 0.000002  \\
			Rewriter vs OR-Tools &       & 181   & 284   & 1.000000  &       & 53    & 412   & 0.000053  \\
			Rewriter vs AM(sampling) &       & 29    & 436   & 1.000000  &       & 465   & 0     & 0.000002  \\
			Rewriter vs AM(greedy) &       & 24    & 441   & 1.000000  &       & 465   & 0     & 0.000002  \\
			ASWTAM vs Rewriter &       & 434   & 31    & 0.000033  &       & 3     & 462   & 0.000002  \\
			ASWTAM vs ACO &       & 465   & 0     & 0.000002  &       & 452   & 13    & 0.000006  \\
			ASWTAM vs AM(greedy) &       & 285   & 150   & 0.130731  &       & 464   & 1     & 0.000002  \\
			ASWTAM vs OR-Tools &       & 316   & 149   & 0.084035  &       & 0     & 465   & 0.000002  \\
			AM(sampling) vs AM(greedy) &       & 219   & 246   & 1.000000  &       & 443   & 22    & 0.000014  \\
			AM(greedy) vs ACO  &       & 465   & 0     & 0.000002  &       & 15    & 450   & 0.000012  \\
			AM(greedy) vs TS &       & 393   & 72    & 0.000928  &       & 0     & 465   & 0.000002  \\
			AM(greedy) vs OR-Tools  &       & 313   & 152   & 0.095706  &       & 0     & 465   & 0.000002  \\
			\bottomrule[2pt]
		\end{tabular}%
		\label{tab:6}%
	}
\end{table}%
\vspace{0.5cm}
\begin{table*}[htbp]
	\centering\vspace{0.5cm}
	\caption{Comparison of average value and solving time (in seconds) of different algorithms on dataset 1.}
	\resizebox{\textwidth}{2.6cm}{
	\begin{tabular}{m{5.565em}<{\centering}ccccccccccc}
		\toprule[2pt]
		\multirow{2}[4]{*}{Baseline} & \multicolumn{3}{m{12.57em}<{\centering}}{VRP20, Cap30} &       & \multicolumn{3}{m{12.57em}<{\centering}}{VRP50, Cap40} &       & \multicolumn{3}{m{12.57em}<{\centering}}{VRP100, Cap50} \\
		\cmidrule{2-4}\cmidrule{6-8}\cmidrule{10-12}    \multicolumn{1}{c}{} & \multicolumn{1}{m{4.19em}<{\centering}}{mean} & \multicolumn{1}{m{4.19em}<{\centering}}{std} & \multicolumn{1}{m{4.19em}<{\centering}}{time} &       & \multicolumn{1}{m{4.19em}<{\centering}}{mean} & \multicolumn{1}{m{4.19em}<{\centering}}{std} & \multicolumn{1}{m{4.19em}<{\centering}}{time} &       & \multicolumn{1}{m{4.19em}<{\centering}}{mean} & \multicolumn{1}{m{4.19em}<{\centering}}{std} & \multicolumn{1}{m{4.19em}<{\centering}}{time} \\
		\midrule
		Gurobi & \textbf{5.74} & \textbf{0.62}  & 1800  &       & \multicolumn{3}{m{12.63em}<{\centering}}{-} &       & \multicolumn{3}{m{12.63em}<{\centering}}{-} \\
		OR-Tools & 	6.12  & 	1.06  & 	1.25  &       & 	10.55 &	 1.56  &	2.32 &       & 	16.55 & 	2.06  & 	3.12 \\
		\midrule
		TS    & 6.27  & 0.69  & 23.57 &       & 11.4  & 0.96  & 54.53 &       & 18.61 & 1.86  & 113.4 \\
		ACO    & 12.29 & 1.38  & 7.42  &       & 19.86 & 1.97  & 34.81 &       & 37.03 & 4.57  & 126.6 \\
		LNS  &6.48  & 	0.96 & 	4.96  &       & 	12.85 &	 1.62  &	94.32 &       & 	18.86 & 	2.03 & 	771.6 \\
		\midrule
		AM(greedy) & 6.41  & 0.83  & \textbf{0.25} &       & 10.79 & 0.82  & \textbf{0.53} &       & 16.66 & 1.78  & 0.93 \\
		AM(sampling) & 6.42  & 0.83  & 0.3   &       & 10.9  & 0.92  & \textbf{0.53} &       & 19.35 & 1.65  & \textbf{0.91} \\
		ASWTAM & 6.36 & 0.67 & 1.41  &       & 11.03 & \textbf{0.76} & 1.65  &       &16.36 & \textbf{1.58} & 1.82 \\
		Rewriter & 6.83  & 0.92  & 2.15  &       & 12.45 & 1.03  & 5.14  &       & 19.98 & 1.9   & 10.56 \\
		L2I   & 6.1   & 0.73  & 5.93  &       & \textbf{10.34} & 0.86  & 13.33 &       & \textbf{16.11} & 1.8   & 25.25 \\
		\bottomrule[2pt]
	\end{tabular}%
	\label{tab:4}%
}
\end{table*}%
\begin{table*}[htbp]
	\centering\vspace{0.5cm}
	\caption{Comparing average value and solving time (in seconds) of different algorithms on data set 2.}
	\resizebox{\textwidth}{2.6cm}{
	\begin{tabular}{m{5.565em}<{\centering}ccccccccccc}
		\toprule[2pt]
		\multirow{2}[4]{*}{Baseline} & \multicolumn{3}{m{12.57em}<{\centering}}{VRP20, Cap30} &       & \multicolumn{3}{m{12.57em}<{\centering}}{VRP50, Cap40} &       & \multicolumn{3}{m{12.57em}<{\centering}}{VRP100, Cap50} \\
		\cmidrule{2-4}\cmidrule{6-8}\cmidrule{10-12}    \multicolumn{1}{c}{} & \multicolumn{1}{m{4.19em}<{\centering}}{mean} & \multicolumn{1}{m{4.19em}<{\centering}}{std} & \multicolumn{1}{m{4.19em}<{\centering}}{time} &       & \multicolumn{1}{m{4.19em}<{\centering}}{mean} & \multicolumn{1}{m{4.19em}<{\centering}}{std} & \multicolumn{1}{m{4.19em}<{\centering}}{time} &       & \multicolumn{1}{m{4.19em}<{\centering}}{mean} & \multicolumn{1}{m{4.19em}<{\centering}}{std} & \multicolumn{1}{m{4.19em}<{\centering}}{time} \\
		\midrule
		Gurobi & \textbf{20.85} & 5.47  &1800  &       & \multicolumn{3}{m{12.57em}<{\centering}}-&       & \multicolumn{3}{m{12.57em}<{\centering}}- \\
		OR-Tools& 21.83  & 	5.58 & 	\textbf{1.03}  &       & 	38.16 &	 9.01 &	2.12  &       & 	58.28 & 	12.31 & 	2.56 \\
		\midrule
		TS & 	22.51 & 5.73 & 58.86 &       & 38.47 & 9.03&	76.26 &       & 	63.85 & 12.38 & 113.93 \\
		ACO     &35.13 &8.61  &23.51 &       & 87.03 &13.99 & 45.15 &       & 165.79 & 13.95&79.08 \\
		LNS & 	21.09  & 	5.80  & 	20.15  &       & 41.85&	 10.56 &	103.84  &       & 	62.12 & 	13.51  & 	804.10 \\
		\midrule
		AM(greedy)& 40.33& 13.9 &0.42 &       & 149.17 & 73.03& \textbf{1.03} &       &241.25& 138.26 &0.98\\
		AM(sampling) & 37.83 & 9.54 & \textbf{0.35} &       &95.23 & 21.27 & 0.62  &       & 169.13 & 30.31 &\textbf{0.8} \\
		ASWTAM& 28.07 & 7.54  & 1.41  &       &58.02 & 16.49 & 1.94 &       & 110.79 & 22.85 & 2.2\\
		Rewriter &24.45 &6.63 &2.26 &       & 44.05 & 9.95 &5.83  &       &69.46 & 12.71 & 12.93 \\
		L2I   & 20.88 &\textbf{5.41} &6.01 &       & \textbf{35.43} & \textbf{8.75} & 13.6  &       &\textbf{54.44}& 12.35 &25.13 \\
		\bottomrule[2pt]
	\end{tabular}%
	\label{tab:5}%
}
\end{table*}%

\subsubsection{Comparison and discussion of Part \Rmnum{3}}
\begin{figure}[htb]
	\flushleft
	\vspace{-0.5cm}
		\includegraphics[width=3.5in]{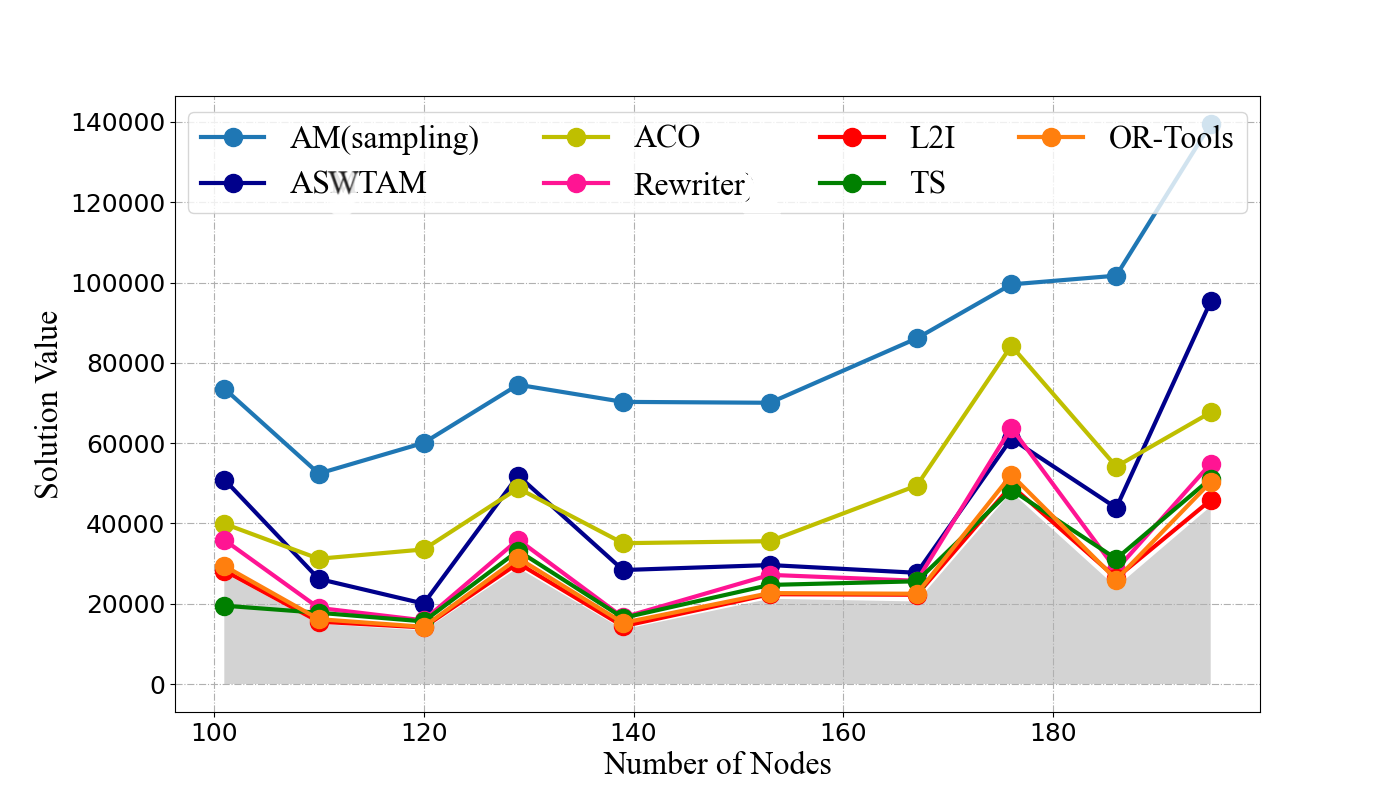}
	\caption{Comparation of solution value of different algorithms on dataset 3.}\label{Fig:11}
\end{figure}

We trained LBO models on VRP20 and tested them on set 3 to verify the scalability of the LBO models, and we depicted the performance of the experimental algorithms in Table~\ref{tab:8} and Fig.~\ref{Fig:11}. We selected only AM with a sampling search policy for testing.

Note that the solving time of all the algorithms increase as the scale of the problem increased, but the solving time of LBO is always significantly less than that of ACO and TS. The most time-consuming L2I is 33.08 s in instances with 195 nodes, whereas TS requires 319.98 s; the longest computational time of the end-to-end model is \textbf{less than 7 s}, which is significantly less than that of OR-Tools. Second, L2I has the best scalability on benchmarks, and the gap between L2I and the optimal does \textbf{not exceed 0.1}. However, L2I is still inferior to TS in X-n101-k25 and X-n186-k15. Moreover, as shown in Table~\ref{tab:8} and Fig.~\ref{Fig:11}, models of end-to-end approaches are more unstable than the step-by-step approaches for large-scale problems. The maximum gap of AM is up to 5.99, but the minimum gap is only 0.83. However, compared with AM, the ASWTAM exhibits significant improvement, and it is better than Rewriter in X-n186-k15. In the other instances, the solutions of ASWTAM are always better than ACO and close to Rewriter. This indicates that end-to-end approaches are promising, and further research is required to achieve better improvement.
\vspace{-0.5cm}

~\
\subsubsection{Experiments conclusion}To demonstrate the effectiveness of the LBO models, we tested AM, ASWTAM, Rewriter, and L2I on three datasets and compared the LBO approaches with ACO, TS, LNS, Gurobi, and OR-Tools. According to the results of the three experiments, the following conclusions can be drawn: (a) step-by-step approaches have faster convergence and better generalization than end-to-end approaches, but they require more computational time; (b) end-to-end approaches frequently have less computational time both during training and testing, but they require more computational resources; (c) both approaches have limitations in the scale of problems and their performance is affected by data distribution; (d) compared with conventional algorithms, LBO approaches require further improvements.
\section{Conclusion and research trend}\label{sec6}
The LBO algorithms have been successfully applied to a series of optimization problems, and studies on using these algorithms to solve the VRP can be divided into two types: end-to-end and step-by-step approaches. Exhaustive experiments demonstrate that step-by-step approaches have strong scalability but are not suitable for problems requiring solving time, whereas end-to-end approaches can rapidly solve problems but are more dependent on data distribution.

Using the LBO algorithms to solve the VRP is still under research. There are several challenges in the LBO algorithms need to be settled in the future and we suggest several potential research directions of applying the LBO algorithms in the VRP from these limitations.
\begin{itemize}
\item[1)]\textbf{Solving sizable or more complex VRPs based on the decomposition framework.} In 2014, Yao and Liu \cite{ref68} had proposed that scaling up learning algorithms is an important problem. LBO models have difficulty in solving combinatorial optimization problems with high complexity or large scale, because these problems result in a curse of dimensionality and easy to overfit \cite{ref69}. Li \emph{et al.} \cite{ref130} proposed to decompose a multi-objective VRP into a set of subproblems, and their results demonstrated that their framework is better than NSGA-II in problems with five objectives; Fu \emph{et al.} \cite{ref137} also decomposed a sizable TSP into multiple small-scale subproblems and solved each of them using the SL model, and they can effectively solve the TSP with 10000 nodes. Hence, decomposing complex or sizable problems to multiple simple subproblems and solving them seems to be a feasible approach.
\item[2)]\textbf{Combining with conventional algorithms to improve the generality of LBO models.} Nickel \emph{et al.} \cite{ref76} illustrated that the relation learned by an LBO model in knowledge graphs only makes sense when applied to entities of the right type. We also observed in our experiments, LBO models are affected by data distribution, while combining heuristics as search operators can improve the generality of the LBO algorithms. Hence, considering different forms of combination might be beneficial to guaranteeing a better generalization of LBO models. Using a solver to post-process solutions of LBO models \cite{ref108} or using solutions generated by heuristic algorithms to pretrain LBO models \cite{ref140} are all good research directions.
\item[3)]\textbf{Embedding other algorithms to increase training efficiency of LBO models.}  Butler \emph{et al.} \cite{ref70} indicated that the data limitations of the LBO algorithms must be solved. The LBO algorithms frequently require sufficient data as a training set, but it is difficult to obtain sufficient raw data of combinatorial optimization problems in real world.  What's more, the value function of learning models is randomly initialized, and models would select a random action with a certain probability to balance exploration and exploitation. LBO models might require a large amount of time and data to train but still converge to a suboptimal solution. Hence, embedding other algorithms to accelerate convergence of LBO model is necessary. For example, using other algorithms computes the initial value function \cite{Ibrahim, ref162} or generates the prior knowledge for the next action \cite{ref139}.
\item[4)]\textbf{Building a generic framework for the LBO algorithms to solve VRPs.} Cappart \emph{et al.} \cite{ref77} proposed to build a generic modeling framework for LBO algorithms as a new direction; Wagstaff \cite{ref78} also indicated that the LBO algorithms have not yet be such matured that researchers from other areas can simply apply them.  Although many LBO algorithms have been proposed to solve different VRPs, encapsulating different LBO algorithms as a solver is
still difficult for researchers. Many technical difficulties require to be solved and the most critical step is to define a generic modeling framework to provide upper-application with uniform interface. 
\end{itemize}
\begin{sidewaystable}[]
	\begin{center}
		\caption{The performance of different algorithms on data set 3. The gap is between experimental algorithms and the optimal solution given by the instance. The solving time is computed in seconds.}
		\centering
		\resizebox{\textwidth}{2.6cm}{
			\begin{tabular}{m{5em}<{\centering}cccccccccccccccccccccccccccc}
				\toprule[2pt]
				\multirow{2}[4]{*}{Instance} & \multicolumn{1}{c}{\multirow{2}[4]{*}{Optimal}} & \multicolumn{3}{m{5.94em}<{\centering}}{ASWTAM} &       & \multicolumn{3}{m{6.505em}<{\centering}}{Rewriter} &       & \multicolumn{3}{m{5.505em}<{\centering}}{AM } &       & \multicolumn{3}{m{6.19em}<{\centering}}{L2I} &       & \multicolumn{3}{m{4.815em}<{\centering}}{ACO} &       & \multicolumn{3}{m{6.005em}<{\centering}}{TS} &       & \multicolumn{3}{m{5.69em}<{\centering}}{OR-Tools} \\
				\cmidrule{3-5}\cmidrule{7-9}\cmidrule{11-13}\cmidrule{15-17}\cmidrule{19-21}\cmidrule{23-25}\cmidrule{27-29}    \multicolumn{1}{c}{} &       & \multicolumn{1}{m{2.19em}<{\centering}}{Value} & \multicolumn{1}{m{2.125em}<{\centering}}{Time} & \multicolumn{1}{m{1.625em}<{\centering}}{Gap} &       & \multicolumn{1}{m{2.375em}<{\centering}}{Value} & \multicolumn{1}{m{2.19em}<{\centering}}{Time} & \multicolumn{1}{m{1.94em}<{\centering}}{Gap} &       & \multicolumn{1}{m{2.315em}<{\centering}}{Value} & \multicolumn{1}{m{1.815em}<{\centering}}{Time} & \multicolumn{1}{m{1.375em}<{\centering}}{Gap } &       & \multicolumn{1}{m{2.25em}<{\centering}}{Value} & \multicolumn{1}{m{2.25em}<{\centering}}{Time} & \multicolumn{1}{m{1.69em}<{\centering}}{Gap } &       & \multicolumn{1}{m{2.065em}<{\centering}}{Value} & \multicolumn{1}{m{1.625em}<{\centering}}{Time} & \multicolumn{1}{m{1.125em}<{\centering}}{Gap } &       & \multicolumn{1}{m{2.065em}<{\centering}}{Value} & \multicolumn{1}{m{1.94em}<{\centering}}{Time} & \multicolumn{1}{m{2em}<{\centering}}{Gap } &       & \multicolumn{1}{m{1.94em}<{\centering}}{Value} & \multicolumn{1}{m{1.875em}<{\centering}}{Time} & \multicolumn{1}{m{1.875em}<{\centering}}{Gap} \\
				\midrule
				X-n101-k25 & 27591 & 50934 & 4.81  & 0.84 &       & 35879 & 11.24 & 0.30  &       & 73578 & 3.95  & 0.83  &       & 28247 & 24.23 & 0.02  &       & 39913 & 61.21 & 0.5   &       & 19504 & 188.89 & \textbf{-0.29} &       & 29405 & 0.53  & 0.06 \\
				X-n110-k13 & 14971 & 26176 & 4.83  & 0.75  &       & 18948 & 11.13 & 0.26  &       & 52410 & 3.45  & 2.91  &       & 15526 & 24.1  & 0.02  &       & 31192 & 50.03 & 1.1   &       & 17745 & 211   & 0.18  &       & 16149& 3.43  &0.08 \\
				X-n120-k6  & 13332 & 20020 & 4.32 & 0.5  &       & 15848 & 10.97 & 0.23  &       & 60056 & 3.89  & 5.99  &       & 14095 & 25.53 & 0.05  &       & 33517 & 57.26 & 1.5   &       & 15591 & 208.53 & 0.17  &       &14243 & 4.76  & 0.07 \\
				X-n129-k18 & 28940 & 51904& 4.68  &0.79&       & 35966 & 12.49 & 0.17  &       & 74536 & 4.34  & 2.55  &       & 30093 & 30.19 & 0.05  &       & 48745 & 51.47 & 0.7   &       & 33196 & 309.08 & 0.15  &       &31362 & 10.02 & 0.08\\
				X-n139-k10 & 13590 & 28382& 5.48  &1.08  &       & 16579 & 12.54 & 0.23  &       & 70266 & 4.31  & 4.39  &       & 14394 & 28.06 & 0.04  &       & 35071 & 70.86 & 1.6   &       & 16479 & 315   & 0.21  &       & 15223 & 11.28 & 0.12 \\
				X-n153-k22 & 21220 & 29631 & 5.15  & 0.4   &       & 27183 & 14.61 & 0.28  &       & 70046 & 4.12  & 2.09  &       & 22364 & 24.06 & 0.06  &       & 35577 & 94.01 & 0.7   &       & 24678 & 444.8 & 0.16  &       & 22650 & 23.4 & 0.07 \\
				X-n176-k26 & 20557 & 27697 & 5.26 & 0.35 &       & 25716 & 15.08 & 0.25  &       & 86097 & 4.69  & 3.18  &       & 22225 & 40.31 & 0.08  &       & 49431 & 85.54 & 1.4   &       & 25582 & 348.16 & 0.24  &       & 22477& 22.85 &0.09 \\
				X-n186-k15 & 47812 &61107 & 6.1   & 0.28  &       & 63848 & 17.47 & 0.33  &       & 99533 & 5.53  & 1.08  &       & 49412 & 31.93 & 0.03  &       & 84249 & 142.9 & 0.8   &       & 48386 & 341.64 & 0.01  &       & 52111 & 44.38 & 0.09 \\
				X-n195-k51 & 24145 & 43872 & 6.19  & 0.82 &       & 28430 & 16.96 & 0.18  &       & 101705 & 6     & 3.21  &       & 26402 & 33.08 & 0.09  &       & 54150 & 136.3 & 1.2   &       & 31176 & 319.98 & 0.29  &       & 26017 & 54.37 &0.07 \\
				\bottomrule[2pt]
			\end{tabular}%
			\label{tab:8}%
		}
	\end{center}
\end{sidewaystable}%

\ifCLASSOPTIONcaptionsoff
\newpage
\fi
%
\bibliographystyle{IEEEtran}
\bibliography{LBOVRP}
\vspace{-0.5cm}
\begin{IEEEbiography}[{\includegraphics[width=0.8in,height=1.1in, clip, keepaspectratio]{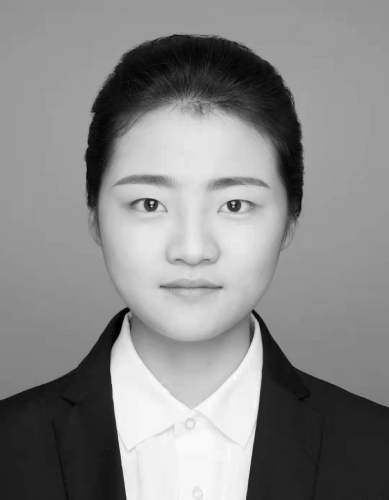}}] {Bingjie Li} received her B.E. degree from Central South University, China, in 2019. Currently, she is working toward her M.E. degree at the School of Traffic \& Transportation Engineering, Central South University.  Her research interests include machine learning and path planning.
\end{IEEEbiography}
\vspace{-0.5cm}
\begin{IEEEbiography}[{\includegraphics[width=0.9in,height=1.3in, clip, keepaspectratio]{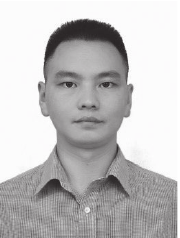}}] {Guohua Wu} received the B.S. degree in Information Systems and Ph.D degree in Operations Research from National University of Defense Technology, China, in 2008 and 2014, respectively. During 2012 and 2014, he was a visiting Ph.D student at University of Alberta, Edmonton, Canada. He is currently a Professor at the School of Traffic and Transportation Engineering, Central South University, Changsha, China.
	His current research interests include Planning and Scheduling, Computational Intelligence and Machine Learning. He has authored more than 80 referred papers including those published in IEEE TCYB, IEEE TSMCA and IEEE TITS. He serves as an Associate Editor of Swarm and Evolutionary Computation Journal, an editorial board member of International Journal of Bio-Inspired Computation, and a Guest Editor of Information Sciences and Memetic Computing. He is a regular reviewer of more than 20 journals including IEEE TEVC, IEEE TCYB, IEEE TSMCA and Information Sciences.
\end{IEEEbiography}
\begin{IEEEbiography}[{\includegraphics[width=0.8in,height=1.1in, clip, keepaspectratio]{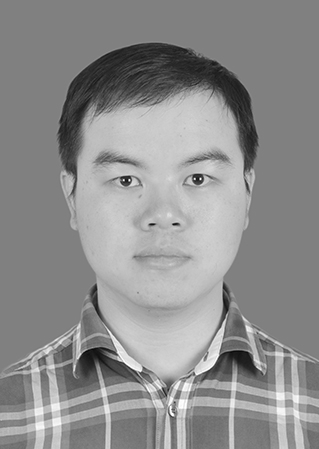}}] {Yongming He} received the B.S. degree in logistics engineering from Chang’an University, Xi’an, China, in 2014. 
	
	He is currently pursuing the Ph.D. degree in management science and engineering with the College of Systems Engineering, National University of Defense Technology, Changsha, China. He was a visiting Ph.D. student with the University of Alberta, Edmonton, AB, Canada, from November 2018 to November 2019. His research interests include operations research, artificial intelligence, intelligent decision, task scheduling, and planning.
\end{IEEEbiography}
\vspace{-1cm}
\begin{IEEEbiography}[{\includegraphics[width=0.8in,height=1.1in, clip, keepaspectratio]{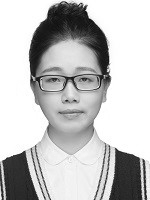}}] {Mingfeng Fan}  received the B.S. degree in Transport Equipment and Control Engineering from Central South University, Changsha, China, in 2019. She is currently pursuing the Ph.D. degree in Traffic and Transportation Engineering with Central South University, Changsha, China. Her research interests include machine learning and UAV path planning.
\end{IEEEbiography}
\vspace{-1cm}
\begin{IEEEbiography}[{\includegraphics[width=1.0in,height=1.5in, clip, keepaspectratio]{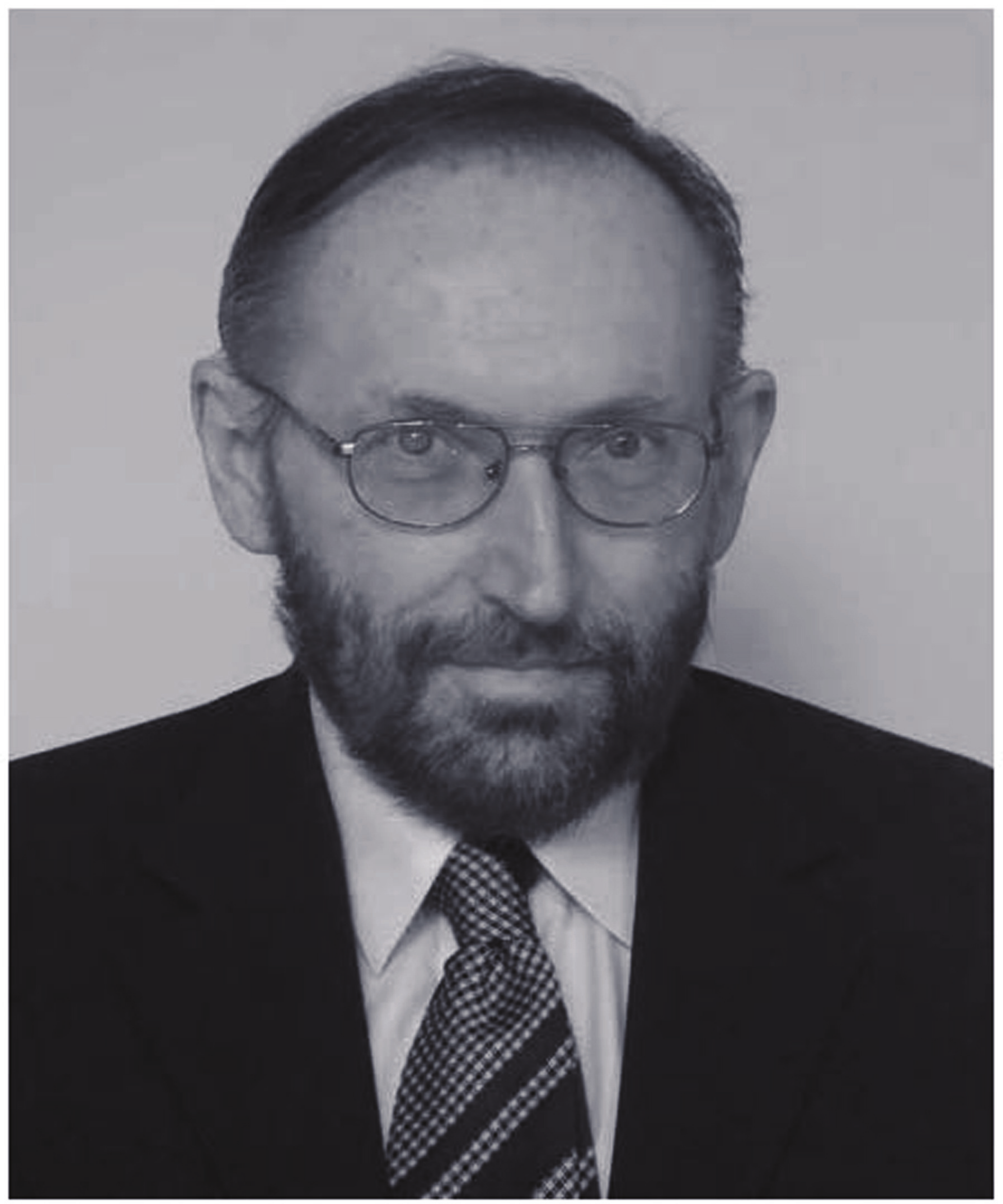}}] {Witold Pedrycz} is a Professor and the Canada Research Chair (CRC-Computational Intelligence) with the Department of Electrical and Computer Engineering, University of Alberta, Edmonton, AB, Canada, and also with the Department of Electrical and Computer Engineering, Faculty of Engineering, King Abdulaziz University, Jeddah, Saudi Arabia. He is also with the Systems Research Institute, Polish Academy of Sciences, Warsaw, Poland. In 2012 he was elected a Fellow of the Royal Society of Canada.
\end{IEEEbiography}
\onecolumn
\begin{landscape}
	\setlength{\tabcolsep}{2.6pt}	
	\setlength\LTleft{-0.6in}
	\small
	\begin{longtable}{lm{8.90em}<{\centering}m{10.19em}<{\centering}m{3.19em}<{\centering}m{3.19em}<{\centering}m{3.19em}<{\centering}m{3.19em}<{\centering}m{1.19em}<{\centering}m{1.19em}<{\centering}m{1.19em}<{\centering}m{2.19em}<{\centering}m{2.19em}<{\centering}m{3.19em}<{\centering}m{2.19em}<{\centering}m{2.19em}<{\centering}m{2.19em}<{\centering}m{2.19em}<{\centering}}
		\caption{Reviewed papers about learning-based optimization algorithms.}\\
		\endfirsthead
		\toprule[2pt]
		\multicolumn{1}{c}{\multirow{2}{*}{year}} &	\multicolumn{1}{c}{\multirow{2}{*}{Reference}}  & \multirow{2}{*}{Problem} & \multicolumn{2}{c}{Reference types} & \multicolumn{2}{c}{Approaches} & \multicolumn{3}{c}{Objective} & \multicolumn{2}{c}{Benchmarks} & \multicolumn{2}{c}{Learning manner} & \multicolumn{3}{c}{Baselines} \\
		\cmidrule{4-17}          & \multicolumn{1}{c}{} & \multicolumn{1}{c}{} & \multicolumn{1}{c}{Conference} & \multicolumn{1}{c}{Journal}& \multicolumn{1}{c}{E2E} & \multicolumn{1}{c}{SbS}  & \multicolumn{1}{c}{Length} & \multicolumn{1}{c}{Time} & \multicolumn{1}{c}{Cost} & \multicolumn{1}{c}{Simulation} &  \multicolumn{1}{c}{Real data} & \multicolumn{1}{c}{SL} & \multicolumn{1}{c}{RL}   & \multicolumn{1}{c}{Solver} & \multicolumn{1}{c}{Heuristics} & \multicolumn{1}{c}{LBO} \\
		\midrule
		& \% of papers & \multicolumn{1}{c}{} & \multicolumn{1}{c}{32.90}\% & \multicolumn{1}{c}{67.10\%} & \multicolumn{1}{c}{65.70\%} & \multicolumn{1}{c}{34.30\%} & \multicolumn{1}{c}{75.70\%} & \multicolumn{1}{c}{8.60\%} & \multicolumn{1}{c}{21.40\%} & \multicolumn{1}{c}{51.40\%} & \multicolumn{1}{c}{60.00\%} & \multicolumn{1}{c}{32.90\%} & \multicolumn{1}{c}{70.00\%} & \multicolumn{1}{c}{52.90\%} & \multicolumn{1}{c}{75.70\%} & \multicolumn{1}{c}{45.7\%}\\
		\midrule
		\endhead
		\toprule[2pt]
		\multicolumn{1}{c}{\multirow{2}{*}{year}} &	\multicolumn{1}{c}{\multirow{2}{*}{Reference}}  & \multirow{2}{*}{Problem} & \multicolumn{2}{c}{Reference types} & \multicolumn{2}{c}{Approaches} & \multicolumn{3}{c}{Objective} & \multicolumn{2}{c}{Benchmarks} & \multicolumn{2}{c}{Learning manner} & \multicolumn{3}{c}{Baselines} \\
		\cmidrule{4-17}          & \multicolumn{1}{c}{} & \multicolumn{1}{c}{} & \multicolumn{1}{c}{Conference} & \multicolumn{1}{c}{Journal}& \multicolumn{1}{c}{E2E} & \multicolumn{1}{c}{SbS}  & \multicolumn{1}{c}{Length} & \multicolumn{1}{c}{Time} & \multicolumn{1}{c}{Cost} & \multicolumn{1}{c}{Simulation} &  \multicolumn{1}{c}{Real data} & \multicolumn{1}{c}{SL} & \multicolumn{1}{c}{RL}   & \multicolumn{1}{c}{Solver} & \multicolumn{1}{c}{Heuristics} & \multicolumn{1}{c}{LBO} \\
		\midrule
		& \% of papers & \multicolumn{1}{c}{} & \multicolumn{1}{c}{32.90\%} & \multicolumn{1}{c}{67.10\%} & \multicolumn{1}{c}{65.70\%} & \multicolumn{1}{c}{34.30\%} & \multicolumn{1}{c}{75.70\%} & \multicolumn{1}{c}{8.60\%} & \multicolumn{1}{c}{21.40\%} & \multicolumn{1}{c}{51.40\%} & \multicolumn{1}{c}{60.00\%} & \multicolumn{1}{c}{32.90\%} & \multicolumn{1}{c}{70.00\%} & \multicolumn{1}{c}{52.90\%} & \multicolumn{1}{c}{75.70\%} & \multicolumn{1}{c}{45.7\%}\\
		\midrule
		2007  & Lima \emph{et al.} \cite{ref85} & TSP   & \multicolumn{1}{c}{$\surd$} & \multicolumn{1}{c}{} &       & \multicolumn{1}{c}{$\surd$}     & \multicolumn{1}{c}{$\surd$}     &       &       & \multicolumn{1}{c}{} &\multicolumn{1}{c}{$\surd$}     &       & \multicolumn{1}{c}{$\surd$}     & \multicolumn{1}{c}{} & \multicolumn{1}{c}{$\surd$}     & \multicolumn{1}{c}{} \\
		\multirow{2}{*}{2009} & Liu \emph{et al.} \cite{ref87}  & TSP   &       & \multicolumn{1}{c}{$\surd$}     &       & \multicolumn{1}{c}{$\surd$}      & \multicolumn{1}{c}{$\surd$}      &       &       & \multicolumn{1}{c}{} & \multicolumn{1}{c}{$\surd$}      &       & \multicolumn{1}{c}{$\surd$}     & \multicolumn{1}{c}{$\surd$}     & \multicolumn{1}{c}{$\surd$}      & \multicolumn{1}{c}{} \\
		& Reinaldo \emph{et al.} \cite{ref100}  & TSP   & \multicolumn{1}{c}{$\surd$} & \multicolumn{1}{c}{} &       &\multicolumn{1}{c}{$\surd$}     & \multicolumn{1}{c}{$\surd$}    &       &       & \multicolumn{1}{c}{} & \multicolumn{1}{c}{$\surd$}     &       & \multicolumn{1}{c}{} & \multicolumn{1}{c}{} & \multicolumn{1}{c}{$\surd$}     & \multicolumn{1}{c}{$\surd$} \\
		2010  & Meignan \emph{et al.} \cite{ref112} & CVRP,DVRP &       & \multicolumn{1}{c}{$\surd$}    &       & \multicolumn{1}{c}{$\surd$}    & \multicolumn{1}{c}{} & \multicolumn{1}{c}{$\surd$} &       & \multicolumn{1}{c}{} & \multicolumn{1}{c}{$\surd$}     &       & \multicolumn{1}{c}{$\surd$}    & \multicolumn{1}{c}{} & \multicolumn{1}{c}{$\surd$}     & \multicolumn{1}{c}{} \\
		2014  & Asta and Özcan \cite{ref113} & VRPTW & \multicolumn{1}{c}{$\surd$} & \multicolumn{1}{c}{} &       & \multicolumn{1}{c}{$\surd$}    & \multicolumn{1}{c}{} &       & \multicolumn{1}{c}{$\surd$} & \multicolumn{1}{c}{} & \multicolumn{1}{c}{$\surd$}     & \multicolumn{1}{c}{$\surd$} & \multicolumn{1}{c}{} & \multicolumn{1}{c}{} & \multicolumn{1}{c}{$\surd$}    & \multicolumn{1}{c}{} \\
		2015  & Vinyals \emph{et al.} \cite{ref107} & TSP   &       & \multicolumn{1}{c}{$\surd$}     & \multicolumn{1}{c}{$\surd$} & \multicolumn{1}{c}{} & \multicolumn{1}{c}{$\surd$}     &       &       & \multicolumn{1}{c}{$\surd$}     & \multicolumn{1}{c}{} & \multicolumn{1}{c}{$\surd$} & \multicolumn{1}{c}{} & \multicolumn{1}{c}{$\surd$}     & \multicolumn{1}{c}{} & \multicolumn{1}{c}{} \\
		\multicolumn{1}{c}{\multirow{2}{*}{2016}} & Bello \emph{et al.} \cite{ref127} & TSP   &       & \multicolumn{1}{c}{$\surd$}   & \multicolumn{1}{c}{$\surd$} & \multicolumn{1}{c}{} & \multicolumn{1}{c}{$\surd$}    &       &       & \multicolumn{1}{c}{$\surd$}    & \multicolumn{1}{c}{} &       &\multicolumn{1}{c}{$\surd$}    & \multicolumn{1}{c}{$\surd$}    & \multicolumn{1}{c}{$\surd$}    & \multicolumn{1}{c}{$\surd$} \\
		& Martin \emph{et al.} \cite{ref118} & CVRP  &       & \multicolumn{1}{c}{$\surd$}    &       & \multicolumn{1}{c}{$\surd$}    & \multicolumn{1}{c}{$\surd$}     &       &       & \multicolumn{1}{c}{} & \multicolumn{1}{c}{$\surd$}     &       & \multicolumn{1}{c}{$\surd$}     & \multicolumn{1}{c}{} & \multicolumn{1}{c}{$\surd$}     & \multicolumn{1}{c}{} \\
		\multirow{4}{*}{2017} & Dai \emph{et al.} \cite{ref134} & TSP   &       & \multicolumn{1}{c}{$\surd$}     & \multicolumn{1}{c}{$\surd$} & \multicolumn{1}{c}{} &\multicolumn{1}{c}{$\surd$}     &       &       & \multicolumn{1}{c}{} & \multicolumn{1}{c}{$\surd$}     &       & \multicolumn{1}{c}{$\surd$}     & \multicolumn{1}{c}{} & \multicolumn{1}{c}{$\surd$}    & \multicolumn{1}{c}{} \\
		& Levy and Wolf \cite{ref128} & TSP   & \multicolumn{1}{c}{$\surd$} & \multicolumn{1}{c}{} & \multicolumn{1}{c}{$\surd$} & \multicolumn{1}{c}{} & \multicolumn{1}{c}{$\surd$}    &       &       & \multicolumn{1}{c}{$\surd$}    & \multicolumn{1}{c}{} & \multicolumn{1}{c}{$\surd$} & \multicolumn{1}{c}{} & \multicolumn{1}{c}{$\surd$}     & \multicolumn{1}{c}{} & \multicolumn{1}{c}{$\surd$} \\
		& Cooray and Rupasinghe \cite{ref91} & EMVRP &       &  \multicolumn{1}{c}{$\surd$}    &       &  \multicolumn{1}{c}{$\surd$}     & \multicolumn{1}{c}{} &       & \multicolumn{1}{c} {$\surd$} &  \multicolumn{1}{c}{$\surd$}     & \multicolumn{1}{c}{} & \multicolumn{1}{c} {$\surd$} & \multicolumn{1}{c}{} & \multicolumn{1}{c}{} & \multicolumn{1}{c}{} & \multicolumn{1}{c}{} \\
		& Tyasnurita \emph{et al.} \cite{ref114} & OVRP  & \multicolumn{1}{c}{$\surd$} & \multicolumn{1}{c}{} &       & \multicolumn{1}{c}{$\surd$}     & \multicolumn{1}{c}{} &       & \multicolumn{1}{c}{$\surd$} & \multicolumn{1}{c}{} & \multicolumn{1}{c}{$\surd$}    & \multicolumn{1}{c}{$\surd$} & \multicolumn{1}{c}{} & \multicolumn{1}{c}{} & \multicolumn{1}{c}{$\surd$}    & \multicolumn{1}{c}{} \\
		\multirow{11}{*}{2018} & Deudon  \emph{et al.} \cite{ref129} & TSP   &       & \multicolumn{1}{c}{$\surd$}     & \multicolumn{1}{c}{$\surd$} & \multicolumn{1}{c}{} & \multicolumn{1}{c}{$\surd$}     &       &       &\multicolumn{1}{c}{$\surd$}     & \multicolumn{1}{c}{} &       &\multicolumn{1}{c}{$\surd$}     & \multicolumn{1}{c}{$\surd$}     & \multicolumn{1}{c}{} & \multicolumn{1}{c}{$\surd$}\\
		& Kaempfer and Wolf \cite{ref131} & MTSP  &       & \multicolumn{1}{c}{$\surd$}    & \multicolumn{1}{c}{$\surd$} & \multicolumn{1}{c}{} & \multicolumn{1}{c}{$\surd$}     &       &       & \multicolumn{1}{c}{} & \multicolumn{1}{c}{$\surd$}     & \multicolumn{1}{c}{$\surd$} & \multicolumn{1}{c}{} & \multicolumn{1}{c}{$\surd$}    & \multicolumn{1}{c}{} & \multicolumn{1}{c}{$\surd$} \\
		& Groshev \emph{et al.} \cite{ref140} & TSP   & \multicolumn{1}{c}{$\surd$} & \multicolumn{1}{c}{} & \multicolumn{1}{c}{$\surd$} & \multicolumn{1}{c}{} &\multicolumn{1}{c}{$\surd$}     &       &       &\multicolumn{1}{c}{$\surd$}    & \multicolumn{1}{c}{} & \multicolumn{1}{c}{$\surd$} & \multicolumn{1}{c}{} & \multicolumn{1}{c}{} & \multicolumn{1}{c}{$\surd$}    & \multicolumn{1}{c}{} \\
		& Ottoni \emph{et al.} \cite{ref136}  & TSP,ATSP &       &\multicolumn{1}{c}{$\surd$}     & \multicolumn{1}{c}{$\surd$}  & \multicolumn{1}{c}{} & \multicolumn{1}{c}{$\surd$}    &       &       & \multicolumn{1}{c}{} & \multicolumn{1}{c}{$\surd$}      &       & \multicolumn{1}{c}{$\surd$}      & \multicolumn{1}{c}{} & \multicolumn{1}{c}{} &\multicolumn{1}{c}{$\surd$}  \\
		& Al-Duoli \emph{et al.} \cite{ref92} & CVRP  & \multicolumn{1}{c}{$\surd$} & \multicolumn{1}{c}{} &       &\multicolumn{1}{c}{$\surd$}    & \multicolumn{1}{c}{$\surd$}    &       &       & \multicolumn{1}{c}{} &\multicolumn{1}{c}{$\surd$}     & \multicolumn{1}{c}{$\surd$} & \multicolumn{1}{c}{} & \multicolumn{1}{c}{} &\multicolumn{1}{c}{$\surd$}     & \multicolumn{1}{c}{} \\
		& Alipour \emph{et al.} \cite{ref86} & TSP   &       & \multicolumn{1}{c}{$\surd$}     &       & \multicolumn{1}{c}{$\surd$}     & \multicolumn{1}{c}{$\surd$}     &       &       & \multicolumn{1}{c}{} &\multicolumn{1}{c}{$\surd$}     &       & \multicolumn{1}{c}{$\surd$}     & \multicolumn{1}{c}{} & \multicolumn{1}{c}{$\surd$}      & \multicolumn{1}{c}{} \\
		& Phiboonbanakit \emph{et al.} \cite{ref88} & CVRP  & \multicolumn{1}{c} {$\surd$} & \multicolumn{1}{c}{} &       &  \multicolumn{1}{c}{$\surd$}     &  \multicolumn{1}{c}{$\surd$}     &       & \multicolumn{1}{c} {$\surd$} & \multicolumn{1}{c}{} &  \multicolumn{1}{c}{$\surd$}     &       &  \multicolumn{1}{c}{$\surd$}    & \multicolumn{1}{c}{} &  \multicolumn{1}{c}{$\surd$}    & \multicolumn{1}{c}{} \\
		& Kerschke \emph{et al.} \cite{ref115} & TSP   &       & \multicolumn{1}{c}{$\surd$}    &       & \multicolumn{1}{c}{$\surd$}    & \multicolumn{1}{c}{$\surd$}    &       &       & \multicolumn{1}{c}{} & \multicolumn{1}{c}{$\surd$}     & \multicolumn{1}{c}{$\surd$}& \multicolumn{1}{c}{} & \multicolumn{1}{c}{$\surd$}   & \multicolumn{1}{c}{} & \multicolumn{1}{c}{} \\
		& Nazari  \emph{et al.} \cite{ref145} & CVRP  &       & \multicolumn{1}{c}{$\surd$}     & \multicolumn{1}{c}{$\surd$} & \multicolumn{1}{c}{} & \multicolumn{1}{c}{$\surd$}     &       &       & \multicolumn{1}{c}{$\surd$}   & \multicolumn{1}{c}{} &       & \multicolumn{1}{c}{$\surd$}    &\multicolumn{1}{c} {$\surd$}     & \multicolumn{1}{c}{$\surd$}     & \multicolumn{1}{c}{$\surd$}\\
		& Kool \emph{et al.} \cite{ref74} & CVRP TSP & \multicolumn{1}{c}{$\surd$}  & \multicolumn{1}{c}{} & \multicolumn{1}{c}{$\surd$}  & \multicolumn{1}{c}{} & \multicolumn{1}{c}{$\surd$}     &       &       & \multicolumn{1}{c}{$\surd$}     & \multicolumn{1}{c}{} &       & \multicolumn{1}{c}{$\surd$}      & \multicolumn{1}{c}{$\surd$}     & \multicolumn{1}{c}{$\surd$}     & \multicolumn{1}{c}{$\surd$}  \\
		& Yao \emph{et al.} \cite{ref73} & Multi-objective route planning &       & \multicolumn{1}{c}{$\surd$}      &       & \multicolumn{1}{c}{$\surd$}      & \multicolumn{1}{c}{$\surd$}     &       &       & \multicolumn{1}{c}{} & \multicolumn{1}{c}{$\surd$}      &       & \multicolumn{1}{c}{$\surd$}    & \multicolumn{1}{c}{} & \multicolumn{1}{c}{$\surd$}   & \multicolumn{1}{c}{} \\
		\multirow{2}{*}{} & Yang \emph{et al.} \cite{ref101} & TSP   & \multicolumn{1}{c}{$\surd$} & \multicolumn{1}{c}{} &       & \multicolumn{1}{c}{$\surd$}    &\multicolumn{1}{c}{$\surd$}   &       &       & \multicolumn{1}{c}{} &\multicolumn{1}{c}{$\surd$}     & \multicolumn{1}{c}{$\surd$} & \multicolumn{1}{c}{} & \multicolumn{1}{c}{} &\multicolumn{1}{c}{$\surd$}    & \multicolumn{1}{c}{} \\
		& Chen and Tian \cite{ref103} & CVRP  &       & \multicolumn{1}{c}{$\surd$}     &       & \multicolumn{1}{c}{$\surd$}     & \multicolumn{1}{c}{$\surd$}   &       &       & \multicolumn{1}{c}{$\surd$}   & \multicolumn{1}{c}{} &       & \multicolumn{1}{c}{$\surd$}   &\multicolumn{1}{c}{$\surd$}    &\multicolumn{1}{c}{$\surd$}     & \multicolumn{1}{c}{$\surd$}\\
		\multirow{14}{*}{2019} & Joshi \emph{et al.} \cite{ref141}& TSP   &       & \multicolumn{1}{c}{$\surd$}    & \multicolumn{1}{c}{$\surd$}& \multicolumn{1}{c}{} & \multicolumn{1}{c}{$\surd$}    &       &       & \multicolumn{1}{c}{$\surd$}   & \multicolumn{1}{c}{$\surd$}  &       & \multicolumn{1}{c}{} & \multicolumn{1}{c}{} &\multicolumn{1}{c}{$\surd$}    &\multicolumn{1}{c}{$\surd$} \\
		& Prates \emph{et al.} \cite{ref142} & TSP   & \multicolumn{1}{c}{$\surd$} & \multicolumn{1}{c}{} & \multicolumn{1}{c}{$\surd$} & \multicolumn{1}{c}{} &\multicolumn{1}{c}{$\surd$}   &       &       &\multicolumn{1}{c}{$\surd$}   &\multicolumn{1}{c}{$\surd$}   & \multicolumn{1}{c}{$\surd$}& \multicolumn{1}{c}{} & \multicolumn{1}{c}{} & \multicolumn{1}{c}{$\surd$}   & \multicolumn{1}{c}{} \\
		& Kalakanti \emph{et al.} \cite{ref168} & SDVRP, VRPTW & \multicolumn{1}{c}{$\surd$} & \multicolumn{1}{c}{} & \multicolumn{1}{c}{$\surd$} & \multicolumn{1}{c}{} & \multicolumn{1}{c}{$\surd$}     & \multicolumn{1}{c}{$\surd$} &       & \multicolumn{1}{c}{} & \multicolumn{1}{c}{$\surd$}    &       &\multicolumn{1}{c}{$\surd$}   & \multicolumn{1}{c}{} &\multicolumn{1}{c}{$\surd$}     & \multicolumn{1}{c}{} \\
		& Holler \emph{et al.} \cite{ref169} & MDVDRP & \multicolumn{1}{c}{$\surd$}& \multicolumn{1}{c}{} & \multicolumn{1}{c}{$\surd$} & \multicolumn{1}{c}{} &\multicolumn{1}{c}{$\surd$}    &       & \multicolumn{1}{c}{$\surd$} & \multicolumn{1}{c}{} &\multicolumn{1}{c}{$\surd$}     &       & \multicolumn{1}{c}{$\surd$}     & \multicolumn{1}{c}{} & \multicolumn{1}{c}{} & \multicolumn{1}{c}{} \\
		& Mukhutdinov \emph{et al.} \cite{ref162} & packet routing problem &       & \multicolumn{1}{c}{$\surd$}     & \multicolumn{1}{c}{$\surd$}& \multicolumn{1}{c}{} & \multicolumn{1}{c}{$\surd$}     &       &       & \multicolumn{1}{c}{$\surd$}    & \multicolumn{1}{c}{} & \multicolumn{1}{c}{$\surd$} & \multicolumn{1}{c}{$\surd$}    & \multicolumn{1}{c}{} & \multicolumn{1}{c}{} & \multicolumn{1}{c}{} \\
		& Ma \emph{et al.} \cite{ref133} & TSP, TSPTW &       & \multicolumn{1}{c}{$\surd$}       & \multicolumn{1}{c}{$\surd$} & \multicolumn{1}{c}{} & \multicolumn{1}{c}{$\surd$}      &       &       & \multicolumn{1}{c}{$\surd$}    & \multicolumn{1}{c}{} &       & \multicolumn{1}{c}{$\surd$}       & \multicolumn{1}{c}{$\surd$}      & \multicolumn{1}{c}{$\surd$}      & \multicolumn{1}{c}{$\surd$}  \\
		& Li \emph{et al.} \cite{ref158} & Online route planning &       & \multicolumn{1}{c}{$\surd$}     & \multicolumn{1}{c}{$\surd$} & \multicolumn{1}{c}{} & \multicolumn{1}{c}{} & \multicolumn{1}{c}{$\surd$}  &       & \multicolumn{1}{c}{} & \multicolumn{1}{c}{$\surd$}     & \multicolumn{1}{c}{$\surd$}  & \multicolumn{1}{c}{} & \multicolumn{1}{c}{} &\multicolumn{1}{c}{$\surd$}    & \multicolumn{1}{c}{} \\
		& Balaji \emph{et al.} \cite{ref160}& SVRP  &       & \multicolumn{1}{c}{$\surd$}     & \multicolumn{1}{c}{$\surd$}  & \multicolumn{1}{c}{} & \multicolumn{1}{c}{} &       & \multicolumn{1}{c}{$\surd$}  & \multicolumn{1}{c}{} & \multicolumn{1}{c}{$\surd$}     &       &\multicolumn{1}{c}{$\surd$}     &\multicolumn{1}{c}{$\surd$}    & \multicolumn{1}{c}{} & \multicolumn{1}{c}{} \\
		& Yu \emph{et al.} \cite{ref159} & Online route planning &       & \multicolumn{1}{c}{$\surd$}    & \multicolumn{1}{c}{$\surd$} & \multicolumn{1}{c}{} & \multicolumn{1}{c}{} & \multicolumn{1}{c}{$\surd$} &       & \multicolumn{1}{c}{} &\multicolumn{1}{c}{$\surd$}     & \multicolumn{1}{c}{$\surd$} & \multicolumn{1}{c}{} & \multicolumn{1}{c}{$\surd$}   & \multicolumn{1}{c}{$\surd$}    & \multicolumn{1}{c}{$\surd$} \\
		& Vera and Abad \cite{ref148}& CMVRP & \multicolumn{1}{c}{$\surd$}& \multicolumn{1}{c}{} & \multicolumn{1}{c}{$\surd$} & \multicolumn{1}{c}{} & \multicolumn{1}{c}{$\surd$}     &       &       & \multicolumn{1}{c}{$\surd$}     & \multicolumn{1}{c}{} &       & \multicolumn{1}{c}{$\surd$}  & \multicolumn{1}{c}{$\surd$}    & \multicolumn{1}{c}{$\surd$}     & \multicolumn{1}{c}{} \\
		& Peng \emph{et al.} \cite{ref147} & CVRP  & \multicolumn{1}{c}{$\surd$} & \multicolumn{1}{c}{} & \multicolumn{1}{c}{$\surd$} & \multicolumn{1}{c}{} & \multicolumn{1}{c}{$\surd$}    &       &       &\multicolumn{1}{c}{$\surd$}     & \multicolumn{1}{c}{} &       & \multicolumn{1}{c}{$\surd$}    & \multicolumn{1}{c}{$\surd$}   & \multicolumn{1}{c}{$\surd$}    & \multicolumn{1}{c}{$\surd$} \\
		& Hottung and Tierney \cite{ref96} & CVRP, SDVRP &       &  \multicolumn{1}{c}{$\surd$}     &       &  \multicolumn{1}{c}{$\surd$}  & \multicolumn{1}{c}{$\surd$}     &       &       & \multicolumn{1}{c}{} &  \multicolumn{1}{c}{$\surd$}   &       &  \multicolumn{1}{c}{$\surd$}     & \multicolumn{1}{c}{} & \multicolumn{1}{c}{$\surd$}      &  \multicolumn{1}{c}{$\surd$} \\
		& Rodriguez \emph{et al.} \cite{ref117} & VRPTW &       & \multicolumn{1}{c}{$\surd$}    &       & \multicolumn{1}{c}{$\surd$}  & \multicolumn{1}{c}{} &       & \multicolumn{1}{c}{$\surd$} & \multicolumn{1}{c}{} &\multicolumn{1}{c}{$\surd$}    & \multicolumn{1}{c}{$\surd$} & \multicolumn{1}{c}{} & \multicolumn{1}{c}{} & \multicolumn{1}{c}{} & \multicolumn{1}{c}{} \\
		& Lu \emph{et al.} \cite{ref104} & CVRP  & \multicolumn{1}{c}{$\surd$} & \multicolumn{1}{c}{} &       & \multicolumn{1}{c}{$\surd$}   &\multicolumn{1}{c}{$\surd$}    &       &       &\multicolumn{1}{c}{$\surd$}     & \multicolumn{1}{c}{} &       & \multicolumn{1}{c}{$\surd$}    & \multicolumn{1}{c}{$\surd$}   & \multicolumn{1}{c}{$\surd$}    & \multicolumn{1}{c}{$\surd$}\\
		\multirow{25}{*}{2020} & Li \emph{et al.} \cite{ref130} & MOTSP &       & \multicolumn{1}{c}{$\surd$}      & \multicolumn{1}{c}{$\surd$} & \multicolumn{1}{c}{} &\multicolumn{1}{c}{$\surd$}     &       &       & \multicolumn{1}{c}{} & \multicolumn{1}{c}{$\surd$}      &       & \multicolumn{1}{c}{$\surd$}    & \multicolumn{1}{c}{} &\multicolumn{1}{c}{$\surd$}      & \multicolumn{1}{c}{} \\
		& Sultana \emph{et al.} \cite{ref143} & TSP   &       & \multicolumn{1}{c}{$\surd$}      & \multicolumn{1}{c}{$\surd$}   & \multicolumn{1}{c}{} & \multicolumn{1}{c}{$\surd$}     &       &       & \multicolumn{1}{c}{$\surd$}      &\multicolumn{1}{c}{$\surd$}       & \multicolumn{1}{c}{$\surd$}   & \multicolumn{1}{c}{} & \multicolumn{1}{c}{$\surd$}       & \multicolumn{1}{c}{$\surd$}   & \multicolumn{1}{c}{$\surd$}   \\
		& Fu \emph{et al.} \cite{ref137}  & TSP   &       & \multicolumn{1}{c}{$\surd$}      & \multicolumn{1}{c}{$\surd$} & \multicolumn{1}{c}{} & \multicolumn{1}{c}{$\surd$}     &       &       &\multicolumn{1}{c}{$\surd$}    & \multicolumn{1}{c}{} & \multicolumn{1}{c}{$\surd$}  & \multicolumn{1}{c}{$\surd$}      & \multicolumn{1}{c}{$\surd$}    & \multicolumn{1}{c}{$\surd$}  & \multicolumn{1}{c}{$\surd$} \\
		& Zhang\emph{et al.} \cite{ref138}  & TSPTWR &       &\multicolumn{1}{c}{$\surd$}    & \multicolumn{1}{c}{$\surd$}  & \multicolumn{1}{c}{} & \multicolumn{1}{c}{$\surd$}     &       & \multicolumn{1}{c}{$\surd$}  &\multicolumn{1}{c}{$\surd$}     & \multicolumn{1}{c}{$\surd$}     &       & \multicolumn{1}{c}{$\surd$}     & \multicolumn{1}{c}{} & \multicolumn{1}{c}{$\surd$}     & \multicolumn{1}{c}{} \\
		& Xin \emph{et al.} \cite{ref150} & CVRP,TSP &       & \multicolumn{1}{c}{$\surd$}   & \multicolumn{1}{c}{$\surd$}& \multicolumn{1}{c}{} & \multicolumn{1}{c}{$\surd$}   &       &       &\multicolumn{1}{c}{$\surd$} & \multicolumn{1}{c}{} &       &\multicolumn{1}{c}{$\surd$}  &\multicolumn{1}{c}{$\surd$}   & \multicolumn{1}{c}{$\surd$}   &\multicolumn{1}{c}{$\surd$} \\
		& Xing \emph{et al.} \cite{ref139} & CVRP,TSP &       &\multicolumn{1}{c}{$\surd$} & \multicolumn{1}{c}{$\surd$}& \multicolumn{1}{c}{} & \multicolumn{1}{c}{$\surd$}  &       &       &\multicolumn{1}{c}{$\surd$} & \multicolumn{1}{c}{$\surd$} & \multicolumn{1}{c}{$\surd$}&\multicolumn{1}{c}{$\surd$}  &\multicolumn{1}{c}{$\surd$}   & \multicolumn{1}{c}{} &\multicolumn{1}{c}{$\surd$}\\
		& Zhao \emph{et al.} \cite{ref153} & CVRP, VRPTW &       &\multicolumn{1}{c}{$\surd$}    & \multicolumn{1}{c}{$\surd$} & \multicolumn{1}{c}{} & \multicolumn{1}{c}{$\surd$}   &       &       & \multicolumn{1}{c}{$\surd$} & \multicolumn{1}{c}{} &       &\multicolumn{1}{c}{$\surd$}& \multicolumn{1}{c}{$\surd$} & \multicolumn{1}{c}{$\surd$}  &\multicolumn{1}{c}{$\surd$}\\
		& Sultana \emph{et al.} \cite{sultana} & CVRP, TSP &       & \multicolumn{1}{c}{$\surd$}     & \multicolumn{1}{c}{$\surd$}& \multicolumn{1}{c}{} &\multicolumn{1}{c}{$\surd$}   &       &       & \multicolumn{1}{c}{$\surd$}& \multicolumn{1}{c}{} &       & \multicolumn{1}{c}{$\surd$}   &\multicolumn{1}{c}{$\surd$}   & \multicolumn{1}{c}{$\surd$}    & \multicolumn{1}{c}{$\surd$} \\
		& Drori \emph{et al.} \cite{ref154} & CVRP, TSP & \multicolumn{1}{c}{$\surd$}  & \multicolumn{1}{c}{} & \multicolumn{1}{c}{$\surd$} & \multicolumn{1}{c}{} & \multicolumn{1}{c}{$\surd$}      &       &       & \multicolumn{1}{c}{$\surd$}      & \multicolumn{1}{c}{} &       &\multicolumn{1}{c}{$\surd$}   &\multicolumn{1}{c}{$\surd$}     & \multicolumn{1}{c}{$\surd$}  & \multicolumn{1}{c}{$\surd$}  \\
		& Bono \emph{et al.} \cite{ref155}& DS-CVRP, DS-CVRPTW &       & \multicolumn{1}{c}{$\surd$}   & \multicolumn{1}{c}{$\surd$} & \multicolumn{1}{c}{} & \multicolumn{1}{c}{$\surd$}  &       &       & \multicolumn{1}{c}{$\surd$}    & \multicolumn{1}{c}{} &       &\multicolumn{1}{c}{$\surd$} & \multicolumn{1}{c}{$\surd$}   & \multicolumn{1}{c}{$\surd$}   & \multicolumn{1}{c}{$\surd$} \\
		& Cao \emph{et al.} \cite{ref166} & SSP   &       &\multicolumn{1}{c}{$\surd$}      & \multicolumn{1}{c}{$\surd$}  & \multicolumn{1}{c}{} & \multicolumn{1}{c}{} & \multicolumn{1}{c}{$\surd$}  &       & \multicolumn{1}{c}{} & \multicolumn{1}{c}{$\surd$}    &       & \multicolumn{1}{c}{$\surd$}     & \multicolumn{1}{c}{} & \multicolumn{1}{c}{$\surd$}     & \multicolumn{1}{c}{} \\
		& Chen \emph{et al.} \cite{ref167} & CVRP  &       & \multicolumn{1}{c}{$\surd$}    & \multicolumn{1}{c}{$\surd$} & \multicolumn{1}{c}{} & \multicolumn{1}{c}{} &       & \multicolumn{1}{c}{$\surd$} &\multicolumn{1}{c}{$\surd$}     & \multicolumn{1}{c}{} &       &\multicolumn{1}{c}{$\surd$}    & \multicolumn{1}{c}{} & \multicolumn{1}{c}{$\surd$}    & \multicolumn{1}{c}{} \\
		& Ding \emph{et al.} \cite{ref89} & CVRP,TSP & \multicolumn{1}{c}{$\surd$} & \multicolumn{1}{c}{} &       &\multicolumn{1}{c}{$\surd$}   &\multicolumn{1}{c}{$\surd$}    &       &       & \multicolumn{1}{c}{$\surd$}    & \multicolumn{1}{c}{} & \multicolumn{1}{c}{$\surd$}& \multicolumn{1}{c}{} &\multicolumn{1}{c}{$\surd$}   & \multicolumn{1}{c}{$\surd$}  & \multicolumn{1}{c}{} \\
		& Zhao \emph{et al.} \cite{ref116} & TSP   &       & \multicolumn{1}{c}{$\surd$}     &       & \multicolumn{1}{c}{$\surd$}     & \multicolumn{1}{c}{$\surd$}  &       &       & \multicolumn{1}{c}{} &\multicolumn{1}{c}{$\surd$}     & \multicolumn{1}{c}{$\surd$}  & \multicolumn{1}{c}{} & \multicolumn{1}{c}{} & \multicolumn{1}{c}{$\surd$}    & \multicolumn{1}{c}{} \\
		& Vlastelica \emph{et al.} \cite{ref108} & TSP   & \multicolumn{1}{c}{$\surd$} & \multicolumn{1}{c}{} &       &\multicolumn{1}{c}{$\surd$}  & \multicolumn{1}{c}{$\surd$}&       &       & \multicolumn{1}{c}{} & \multicolumn{1}{c}{$\surd$}    & \multicolumn{1}{c}{$\surd$}& \multicolumn{1}{c}{} &\multicolumn{1}{c}{$\surd$}   & \multicolumn{1}{c}{} & \multicolumn{1}{c}{} \\
		& Delarue \emph{et al.} \cite{ref102} & CVRP  &       & \multicolumn{1}{c}{$\surd$}     & \multicolumn{1}{c}{$\surd$}  & \multicolumn{1}{c}{} & \multicolumn{1}{c}{$\surd$}   &       &       & \multicolumn{1}{c}{} &\multicolumn{1}{c}{$\surd$}    &       & \multicolumn{1}{c}{$\surd$}     &\multicolumn{1}{c}{$\surd$}    & \multicolumn{1}{c}{$\surd$}     & \multicolumn{1}{c}{$\surd$}  \\
		& Sheng \emph{et al.} \cite{ref164}  & VRPTRLR &       &\multicolumn{1}{c}{$\surd$}      & \multicolumn{1}{c}{$\surd$} & \multicolumn{1}{c}{} & \multicolumn{1}{c}{} &       & \multicolumn{1}{c}{$\surd$} & \multicolumn{1}{c}{$\surd$}     & \multicolumn{1}{c}{} &       & \multicolumn{1}{c}{$\surd$}      & \multicolumn{1}{c}{} & \multicolumn{1}{c}{$\surd$}    & \multicolumn{1}{c}{} \\
		& Duan \emph{et al.} \cite{Duan} & CVRP  & \multicolumn{1}{c}{$\surd$}   & \multicolumn{1}{c}{} & \multicolumn{1}{c}{$\surd$}   & \multicolumn{1}{c}{} & \multicolumn{1}{c}{} &       & \multicolumn{1}{c}{$\surd$}   & \multicolumn{1}{c}{} & \multicolumn{1}{c}{$\surd$}      & \multicolumn{1}{c}{$\surd$}   & \multicolumn{1}{c}{$\surd$}     & \multicolumn{1}{c}{$\surd$}       & \multicolumn{1}{c}{} & \multicolumn{1}{c}{$\surd$}  \\
		& Zhang \emph{et al.} \cite{ref152} & MVRPSTW &       & \multicolumn{1}{c}{$\surd$}    & \multicolumn{1}{c}{$\surd$}  & \multicolumn{1}{c}{} & \multicolumn{1}{c}{} &       & \multicolumn{1}{c}{$\surd$} & \multicolumn{1}{c}{$\surd$}    & \multicolumn{1}{c}{} &       & \multicolumn{1}{c}{$\surd$}      & \multicolumn{1}{c}{$\surd$}  &\multicolumn{1}{c}{$\surd$} & \multicolumn{1}{c}{} \\
		& Lin \emph{et al.} \cite{ref156} & EVRPTW &       & \multicolumn{1}{c}{$\surd$}    & \multicolumn{1}{c}{$\surd$}  & \multicolumn{1}{c}{} & \multicolumn{1}{c}{$\surd$}     &       &       &\multicolumn{1}{c}{$\surd$} & \multicolumn{1}{c}{} &       &\multicolumn{1}{c}{$\surd$} &\multicolumn{1}{c}{$\surd$}   &\multicolumn{1}{c}{$\surd$}   & \multicolumn{1}{c}{} \\
		& Falkner and Schmidt-Thieme \cite{ref149} & CVRPTW &       & \multicolumn{1}{c}{$\surd$}     & \multicolumn{1}{c}{$\surd$} & \multicolumn{1}{c}{} & \multicolumn{1}{c}{} &       & \multicolumn{1}{c}{$\surd$}  & \multicolumn{1}{c}{} &\multicolumn{1}{c}{$\surd$}    &       & \multicolumn{1}{c}{$\surd$}  & \multicolumn{1}{c}{$\surd$}   & \multicolumn{1}{c}{} &\multicolumn{1}{c}{$\surd$}  \\
		& Chen \emph{et al.} \cite{ref97} & CVRPTW &       & \multicolumn{1}{c}{$\surd$}     &       &\multicolumn{1}{c}{$\surd$}      & \multicolumn{1}{c}{} &       & \multicolumn{1}{c}{$\surd$}  & \multicolumn{1}{c}{} & \multicolumn{1}{c}{$\surd$}      &       & \multicolumn{1}{c}{$\surd$}    & \multicolumn{1}{c}{} &\multicolumn{1}{c}{$\surd$}      & \multicolumn{1}{c}{} \\
		& Joe and Lau \cite{ref72} & DSVRP & \multicolumn{1}{c}{$\surd$}  & \multicolumn{1}{c}{} &       & \multicolumn{1}{c}{$\surd$}  & \multicolumn{1}{c}{} &       & \multicolumn{1}{c}{$\surd$}  & \multicolumn{1}{c}{} & \multicolumn{1}{c}{$\surd$}   &       & \multicolumn{1}{c}{$\surd$}     & \multicolumn{1}{c}{} & \multicolumn{1}{c}{$\surd$}     & \multicolumn{1}{c}{} \\
		& Gao \emph{et al.} \cite{ref98} & CVRP, CVRPTW &       & \multicolumn{1}{c}{$\surd$}   &       & \multicolumn{1}{c}{$\surd$}     & \multicolumn{1}{c}{$\surd$}   &       &       & \multicolumn{1}{c}{$\surd$}     & \multicolumn{1}{c}{} &       &\multicolumn{1}{c}{$\surd$}  & \multicolumn{1}{c}{} & \multicolumn{1}{c}{$\surd$}   &\multicolumn{1}{c}{$\surd$}  \\
		& Costa \emph{et al.} \cite{refda} & TSP   &       & \multicolumn{1}{c}{$\surd$}      &       & \multicolumn{1}{c}{$\surd$}  & \multicolumn{1}{c}{$\surd$}     &       &       & \multicolumn{1}{c}{$\surd$}    & \multicolumn{1}{c}{$\surd$}  &       & \multicolumn{1}{c}{$\surd$}      & \multicolumn{1}{c}{$\surd$}  & \multicolumn{1}{c}{$\surd$}  & \multicolumn{1}{c}{$\surd$} \\
		\multirow{6}{*}{2021} & Le \emph{et al.} \cite{ref132}& TSP   &       & \multicolumn{1}{c}{$\surd$}     & \multicolumn{1}{c}{$\surd$} & \multicolumn{1}{c}{} & \multicolumn{1}{c}{$\surd$}   &       &       &\multicolumn{1}{c}{$\surd$}    & \multicolumn{1}{c}{} &       & \multicolumn{1}{c}{$\surd$}    & \multicolumn{1}{c}{} &\multicolumn{1}{c}{$\surd$}     & \multicolumn{1}{c}{} \\
		& Xin \emph{et al.} \cite{ref151}& CVRP,TSP & \multicolumn{1}{c}{$\surd$}   & \multicolumn{1}{c}{} & \multicolumn{1}{c}{$\surd$}  & \multicolumn{1}{c}{} & \multicolumn{1}{c}{$\surd$}     &       &       & \multicolumn{1}{c}{$\surd$}   & \multicolumn{1}{c}{$\surd$}      &       & \multicolumn{1}{c}{$\surd$}     &\multicolumn{1}{c}{$\surd$}     & \multicolumn{1}{c}{$\surd$}   & \multicolumn{1}{c}{$\surd$}   \\
		& Li \emph{et al.} \cite{ref157} & PDP   &       & \multicolumn{1}{c}{$\surd$}   & \multicolumn{1}{c}{$\surd$}  & \multicolumn{1}{c}{} & \multicolumn{1}{c}{} & \multicolumn{1}{c}{$\surd$}  &       &\multicolumn{1}{c}{$\surd$}     & \multicolumn{1}{c}{} &       & \multicolumn{1}{c}{$\surd$}   &\multicolumn{1}{c}{$\surd$}   & \multicolumn{1}{c}{$\surd$}    &\multicolumn{1}{c}{$\surd$}  \\
		& Lin \emph{et al.} \cite{ref156}  & EVRPTW &       & \multicolumn{1}{c}{$\surd$}    & \multicolumn{1}{c}{$\surd$} & \multicolumn{1}{c}{} & \multicolumn{1}{c}{$\surd$}    &       &       & \multicolumn{1}{c}{} & \multicolumn{1}{c}{$\surd$}    &       & \multicolumn{1}{c}{$\surd$}    & \multicolumn{1}{c}{$\surd$}   &\multicolumn{1}{c}{$\surd$}     & \multicolumn{1}{c}{$\surd$} \\
		& Sun \emph{et al.} \cite{ref90}& TSP   &       & \multicolumn{1}{c}{$\surd$}     &       &\multicolumn{1}{c}{$\surd$}       & \multicolumn{1}{c}{$\surd$}     &       &       & \multicolumn{1}{c}{} &\multicolumn{1}{c}{$\surd$}     & \multicolumn{1}{c}{$\surd$}  & \multicolumn{1}{c}{} & \multicolumn{1}{c}{$\surd$}     & \multicolumn{1}{c}{} & \multicolumn{1}{c}{} \\
		& Wu \emph{et al.} \cite{ref106} & CVRP, TSP &       & \multicolumn{1}{c}{$\surd$}     &       & \multicolumn{1}{c}{$\surd$}      & \multicolumn{1}{c}{$\surd$}  &       &       &\multicolumn{1}{c}{$\surd$}     &\multicolumn{1}{c}{$\surd$}    &       & \multicolumn{1}{c}{$\surd$}   & \multicolumn{1}{c}{$\surd$}   & \multicolumn{1}{c}{$\surd$}   &\multicolumn{1}{c}{$\surd$}   \\
		\toprule[2pt]
		\label{tab:2}%
	\end{longtable}
\end{landscape}
\twocolumn

\end{document}